\documentclass[11pt,a4paper]{article}

\usepackage[utf8]{inputenc}
\usepackage[T1]{fontenc}

\usepackage[a4paper,margin=2.5cm]{geometry}

\usepackage{graphicx}
\usepackage{booktabs}
\usepackage{amsmath}
\usepackage{amsfonts}
\usepackage{float}
\usepackage{subcaption}
\usepackage{caption}
\usepackage{adjustbox}
\usepackage{multirow}
\usepackage{array}
\usepackage[table]{xcolor}
\usepackage{ragged2e}
\usepackage{tikz}
\usetikzlibrary{positioning,arrows.meta,shapes.geometric,calc}

\usepackage{url}
\usepackage[colorlinks=true,linkcolor=blue,citecolor=blue,urlcolor=blue]{hyperref}
\usepackage{cleveref} 

\usepackage{microtype}
\usepackage{nicefrac}
\usepackage{makecell}

\makeatletter
\IfFileExists{soul.sty}{%
  \usepackage{soul}%
  \sethlcolor{yellow}%
  \let\add\hl%
}{%
  \let\add\@firstofone
}
\makeatother

\graphicspath{{media/}}

\definecolor{TableHeader}{HTML}{C8DDE1}
\definecolor{TableRow}{HTML}{EAF4F6}
\definecolor{CellHL}{HTML}{BFDFFF}

\newcolumntype{L}[1]{>{\RaggedRight\arraybackslash}p{#1}}

\usepackage{tikz}
\usetikzlibrary{arrows.meta,calc}
\usepackage{graphicx}
\usepackage{subcaption}

\usepackage{booktabs}
\usepackage{array}
\usepackage[table]{xcolor}
\usepackage{adjustbox} 
\usepackage{ragged2e}  

\definecolor{TableHeader}{HTML}{C8DDE1} 
\definecolor{TableRow}{HTML}{EAF4F6}    
\definecolor{CellHL}{HTML}{BFDFFF}      

\usepackage{adjustbox}
\usepackage{makecell} 
\newcommand{\circnum}[1]{\raisebox{.2ex}{\textcircled{\scriptsize #1}}}

\usepackage{tikz}
\usepackage{xcolor}
\usetikzlibrary{arrows.meta, positioning, calc, shadows}

\usepackage{graphicx}
\usepackage{xcolor}
\usepackage{array}
\usepackage{tikz}
\usetikzlibrary{arrows.meta, positioning, calc, shadows}

\definecolor{pipeBlue}{HTML}{3F7FBF}  
\definecolor{pipeFill}{HTML}{EEF7FF}  
\definecolor{pipeEdge}{HTML}{8DBFE8}  
\definecolor{pipeArrow}{HTML}{5B9BD5} 
\definecolor{relFill}{HTML}{F3FAFF}   
\definecolor{relEdge}{HTML}{A9CCE8}   
\definecolor{relText}{HTML}{2F6FA3}   

\newlength{\gtilew}
\newlength{\tilew}

\begin{document}

\title{SARLO-80: Worldwide Slant SAR Language Optic Dataset at 80 cm Resolution}

\author{Sol\`ene Debuys\`ere\inst{1} \and
Nicolas Trouv\'e\inst{1} \and
Nathan Letheule\inst{1} \and
Elise Colin\inst{1} \and
Georgia Channing\inst{2}}

\title{SARLO-80: Worldwide Slant SAR Language Optic Dataset 80cm}

\author{
Sol\`ene Debuys\`ere$^{1}$ \and
Nicolas Trouv\'e$^{1}$ \and
Nathan Letheule$^{1}$ \and
Elise Colin$^{2}$ \and
Georgia Channing$^{3}$\\[0.5em]
$^{1}$DEMR-ONERA -- The French Aerospace Lab, Universit\'e Paris-Saclay, Palaiseau, France\\
\texttt{\{solene.debuysere,nicolas.trouve,nathan.letheule,elise.colin\}@onera.fr}\\
$^{2}$DTIS-ONERA -- The French Aerospace Lab, Universit\'e Paris-Saclay, Palaiseau, France\\
\texttt{\{solene.debuysere,nicolas.trouve,nathan.letheule,elise.colin\}@onera.fr}\\
$^{3}$Hugging Face, London\\
\texttt{georgia.channing@hugging.co}
}

\date{}

\maketitle
\begin{center}
    \vspace{-0.8em}
    \includegraphics[width=0.55\textwidth]{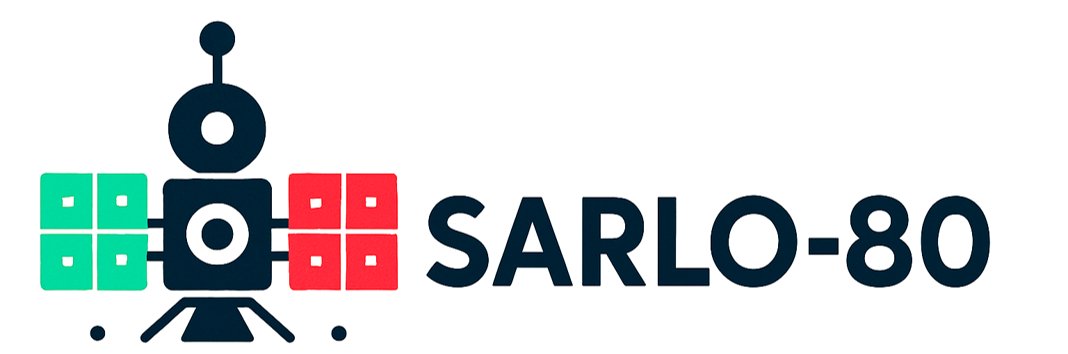}
    \vspace{1em}
\end{center}

\hspace{-0.8em}
\begin{abstract}
Multimodal foundation models have advanced rapidly thanks to large optical benchmarks, but comparable resources for synthetic aperture radar (SAR) remain limited. Existing SAR--optical datasets largely rely on low-resolution, intensity-only Ground Range Detected~(GRD) products and do not preserve complex-valued SAR measurements or native acquisition geometry, which restricts physically grounded multimodal learning. In particular, large-scale public datasets combining very-high-resolution (VHR) SAR SLC, aligned optical imagery, and natural-language descriptions are still lacking.
We present a VHR SAR--optical--text dataset built from open-access Umbra spotlight acquisitions distributed as Sensor Independent Complex Data (SICD). From $\sim$2,500 worldwide scenes (VV/HH, 20\,cm--2\,m native resolution), we standardize all SAR data to an 80\,cm slant-range grid via band-limited FFT resampling and tile the imagery into $1024\times1024$ patches. For each SAR patch, we retrieve a high-resolution optical tile and warp it into the SAR grid using local coordinate correspondences for local pixel-level alignment. We further generate three caption variants (SHORT/MID/LONG) per sample to support vision--language training and evaluation. 
Our dataset contains 119,566 triplets (complex and amplitude slant-range SAR patch, aligned optical patch, natural-language description) covering 257 locations across 72 countries and a broad range of land types and infrastructures. We release fixed train/validation/test splits and the full preprocessing and baseline code to enable reproducible benchmarks for multimodal alignment on conditional generation in native SAR geometry. The dataset is publicly available on the Hugging Face Hub at \mbox{\url{https://huggingface.co/datasets/ONERA/SARLO-80}}.
\end{abstract}

\section{Introduction}
\label{sec:intro}
Synthetic Aperture Radar (SAR) complements optical imagery because it measures the scene structure through radar scattering instead of using visible light. It works day and night and is much less affected by clouds or smoke. Moreover, SAR data are originally provided as complex SLC measurements (amplitude + phase) in the sensor’s native slant-range geometry, where effects such as layover and shadow are visible. Keeping this physical information is valuable, but it also makes SAR harder to use and often requires specific processing and domain knowledge. 

Multimodal learning in remote sensing relies on large and well-curated datasets, yet most existing SAR–optical benchmarks are built from Sentinel-1 Ground Range Detected (GRD) products. These data are intensity-only, ground-projected, and relatively low resolution (typically $\sim$10–30 m), which discards important information contained in complex SAR measurements such as phase and native acquisition geometry. As a result, current datasets provide limited support for learning representations that fully exploit SAR-specific physical properties. In particular, large-scale multimodal resources combining very-high-resolution (VHR) SAR SLC imagery with aligned optical observations and natural-language descriptions remain largely unavailable, and no public vision–language models are currently focused on such data. At the same time, SAR sensing capabilities are rapidly evolving: modern commercial constellations (e.g., Umbra, ICEYE, Capella) can acquire VHR SAR imagery, and recent open-access initiatives are beginning to make these data available to the research community.

In this context, we present our SAR-specific multimodal dataset in native SAR geometry (119,566 samples at 80 cm) built from processing open-access spotlight acquisitions of the Umbra satellite constellation \cite{umbra_open_data}. We start from thousands of complex SAR scenes (SICD) acquired worldwide and processed them by refocusing and resampling at 80 cm. Each scene is then tiled into $1024 \times 1024$ patches in slant-range grid. To form multimodal pairs, we associate each SAR patch with a high-resolution optical image mapped into the SAR slant-range frame via a first-order affine approximation, ensuring pixel-wise correspondence. By preserving the complex-valued SAR signal in native slant-range geometry, the dataset retains the band-limited coherent field required for theoretically sound post-hoc resampling under Shannon–Nyquist, while avoiding detection and reprojection steps that geometrically rewarp layover, foreshortening, and shadowing effects. Then, three captions are generated from the optical modality and post-processed for consistency.

The goal of our baseline experiments is not to establish a new state-of-the-art model, but to provide a simple proof-of-concept illustrating how the proposed dataset can support conditional generation in very-high-resolution SLC SAR. We will release the full preprocessing pipeline, training/validation/testing splits, and evaluation code to facilitate reproducible benchmarking. In general, our dataset provides a foundation for training multimodal SAR-aware models, including vision–language models (VLMs), generative models, and other foundation models for tasks such as super-resolution, semantic understanding, and multimodal reasoning.
  
The remainder of the paper is organized as follows: Section~\cref{sec:related-works} reviews existing SAR datasets and their limitations; Section~\cref{sec:dataset} describes the dataset construction and preprocessing pipeline; Section~\cref{sec:experiments} presents initial experiments for conditional generation in native SAR geometry; and Section~\cref{sec:conclusion} concludes with perspectives and future work. 

\section{Related Works}
\label{sec:related-works}

\paragraph{\textbf{SAR--optical paired datasets.}}
Several open datasets have been proposed to combine SAR and optical imagery for many Earth observation tasks. Early benchmarks such as SEN1-2 \cite{isprs-annals-IV-1-141-2018} pair Sentinel-1 with Sentinel-2 to enable global-scale learning under diverse seasonal conditions, while SEN12MS \cite{schmitt2019sen12mscurateddataset} extends this idea with multi-spectral optical data and additional land-cover information. BigEarthNet-MM \cite{Sumbul_2021} provides a large collection of SAR--optical patches over Europe with consistent labeling, and later regional datasets such as MultiSenGE \cite{isprs-annals-V-3-2022-635-2022} focus on specific areas with tighter acquisition control. More recently, multi-modal resources such as MMEarth \cite{nedungadi2024mmearthexploringmultimodalpretext} and TerraMesh \cite{blumenstiel2025terrameshplanetarymosaicmultimodal} scale up the number of locations and modalities (e.g., DEM, vegetation indices, land-cover layers) to support representation learning across sensors and geographies. Table~\ref{tab:open_sar_optic} summarizes representative datasets and highlights key differences in modality, geometry, and scale.

\begin{table*}[htb]
\centering
\setlength{\tabcolsep}{4pt}
\renewcommand{\arraystretch}{1.15}

\begin{adjustbox}{max width=\textwidth}
\begin{tabular}{%
L{2.6cm}
L{2.2cm}
L{2.2cm}
L{3.2cm}
L{2.1cm}
L{2.5cm}
L{2.6cm}
L{3.1cm}
}
\toprule
\textbf{Dataset} & \textbf{SAR Sensor} & \textbf{SAR Geometry} & \textbf{Coregistered Modality} &
\textbf{SAR Resolution} & \textbf{Patch Size} & \textbf{Sample Count} & \textbf{Coverage} \\
\midrule

\textbf{SEN1-2 (2018)} &
S-1 (C-band) &
GRD &
Sentinel-2 (optical RGB+NIR/MS) &
$\sim$10 m &
256 $\times$ 256 px &
282{,}384 SAR--optical pairs &
Global, all seasons \\

\textbf{SARptical (2018)} &
\textbf{TerraSAR-X (X-band)} &
\textbf{VHR Slant-range (SLC)} &
\textbf{Aerial UltraCAM optical} &
\textbf{$\sim$1 m} &
112 $\times$ 112 px &
10{,}000 SAR--optical pairs &
Dense urban area around Munich, Germany (2009--2013) \\

\textbf{SEN12MS (2019)} &
Sentinel-1 (C-band) &
Dual-pol VV/VH, GRD &
Sentinel-2 MS + MODIS LULC &
10 m &
256 $\times$ 256 px &
180{,}682 triplets &
Global, 4 seasons \\

\textbf{BigEarthNet-MM (2019)} &
S-1 (C-band) &
Dual-pol VV/VH, GRD &
Sentinel-2 MS (12 bands) + CLC labels &
10 m &
120 $\times$ 120 px &
590{,}326 pairs &
10 European countries (2017--2018) \\

\textbf{MultiSenGE (2022)} &
Sentinel-1 (C-band) &
Dual-pol IW GRD &
Sentinel-2 MS, LULC map &
10 m &
256 $\times$ 256 px &
8{,}157 triplets &
Eastern France (2020) \\

\textbf{MMEarth (2024)} &
Sentinel-1 (C-band), 8 bands (VV/VH/ HV/HH asc/desc) &
Map-projected (pixel-level reprojected to S2 10\,m grid) &
12 modalities: S2, DEM + labeled dataset &
10 m &
128 $\times$ 128 px &
1.2M locations &
Global, 14 biomes, 2017--2020 \\

\textbf{TerraMesh (2025)} &
S-1 &
GSD &
S-2, DEM, NDVI, LULC &
10 m &
264 $\times$ 264 px &
\textbf{$\sim$8M co-registered samples} &
Global, multi-year \\

\textbf{BRIGHT (2025)} & 
\textbf{Capella \& Umbra (X-band)} &
GSD &
\textbf{VHR optical} (+ labeled dataset) &
\textbf{0.3--1 m} &
1024 $\times$ 1024 px &
$\sim$4.2--4.5k multimodal image pairs &
14 disaster events in 23 regions worldwide \\

\midrule
\textbf{Our Dataset} &
\textbf{Umbra (X-band, VV/HH)} &
\textbf{VHR Slant-range (SLC)} &
\textbf{HR optical (pixel-aligned with SAR SLC)} &
\textbf{0.8 m} &
\textbf{1024 $\times$ 1024 px} &
$\sim$120{,}000 triplets (SAR complex+amplitude PNG, optical, text)+ metadata (bbox, incidence angle) &
\textbf{Global Coverage} \\

\bottomrule
\end{tabular}
\end{adjustbox}
\caption{Existing open-source SAR--Optical datasets and their characteristics. We highlight VHR and/or Umbra-based resources that are closest to our setting.}
\label{tab:open_sar_optic}
\end{table*}

\paragraph{\textbf{Very-high-resolution (VHR) SAR--optical data}}
Compared to Sentinel-based benchmarks, only a few datasets provide VHR SAR paired with optical imagery. SARptical pairs TerraSAR-X spotlight acquisitions with high-resolution aerial imagery, offering detailed urban content but limited geographic diversity and scale. BRIGHT moves toward modern commercial SAR by combining X-band data (e.g., Capella and Umbra) with VHR optical imagery in a disaster-response context; however, it remains relatively small and focused on a limited set of events.

\paragraph{\textbf{Current limitations}}
Despite their impact, existing open SAR-optical datasets are still constrained compared to optical-only benchmarks, particularly for training SAR-centric foundation models. First, most resources rely on Sentinel-1 \emph{GRD} products, which are intensity-only, ground-projected images and therefore discard complex measurements that encode key SAR physical properties (layover, shadow or foreshortening). Second, many datasets standardize all modalities onto coarse, fixed grids (often around 10 m) by upsampling lower-resolution layers, which simplifies fusion but limits the study of fine-scale structure. Third, patch-based formats dominate with fixed image sizes (e.g., $128\times128$, $256\times256$), and few datasets provide consistent access to the original high-resolution acquisition geometry. Finally, large-scale \emph{multimodal} resources jointly combining VHR SAR, aligned optical imagery, and natural-language descriptions remain largely unavailable. In general, existing open benchmarks also lack temporal and polarimetric configurations that would enable joint analysis of SAR SLC/full-polarimetric data and optical time series.

\paragraph{\textbf{Positioning of our dataset}}
Our dataset addresses these gaps by focusing on (i) \textbf{very-high-resolution SAR} while preserving the \textbf{complex SLC data} in \textbf{native slant-range geometry} (orange plane in Figure \cref{fig:umbra_and_geometry}), enabling principled resampling without geometric reprojection artifacts; (ii) a \textbf{standardized 0.8 m} slant-range representation in a deep learning-ready format (frequency-domain resampling that prioritizes downsampling); and (iii) \textbf{pixel-aligned high-resolution optical imagery} projected onto the SAR grid, complemented with \textbf{natural-language descriptions} at multiple lengths. This design better reflects modern SAR sensing capabilities and enables multimodal learning, retrieval, and foundation-model research that can exploit SAR’s sensor-specific characteristics.

\section{Dataset Creation and methodology}
\label{sec:dataset}
\subsection{Source data: UMBRA Collections}
\begin{figure}[t]
  \centering
  \begin{subfigure}{0.48\linewidth}
    \centering
    \includegraphics[width=\linewidth]{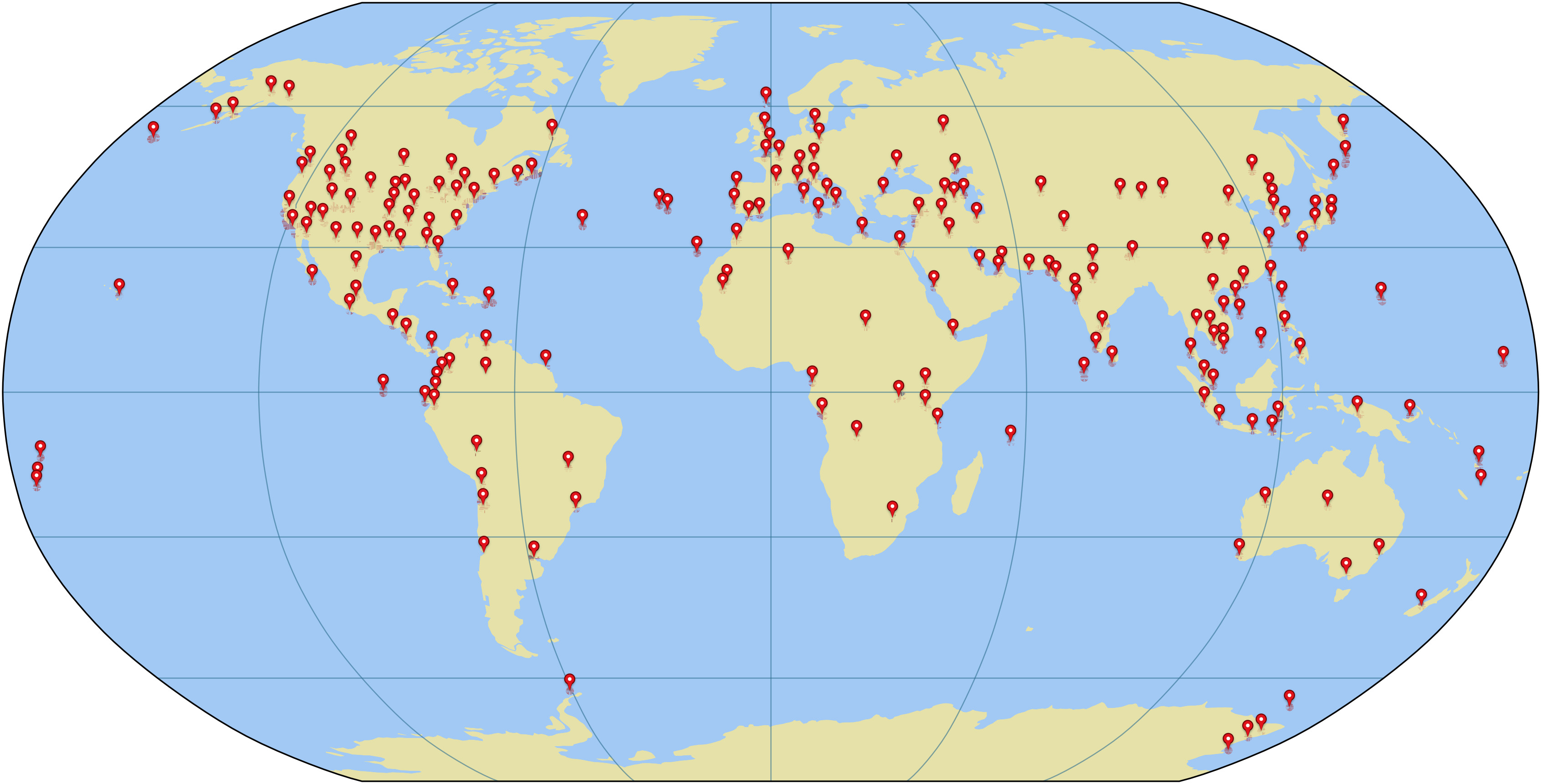}
    \caption{Geographic distribution of the Umbra SICD scenes used to build our dataset.}
    \label{fig:umbra_locations}
  \end{subfigure}
  \hspace{5pt}
  \begin{subfigure}{0.32\linewidth}
    \centering
    \includegraphics[width=\linewidth]{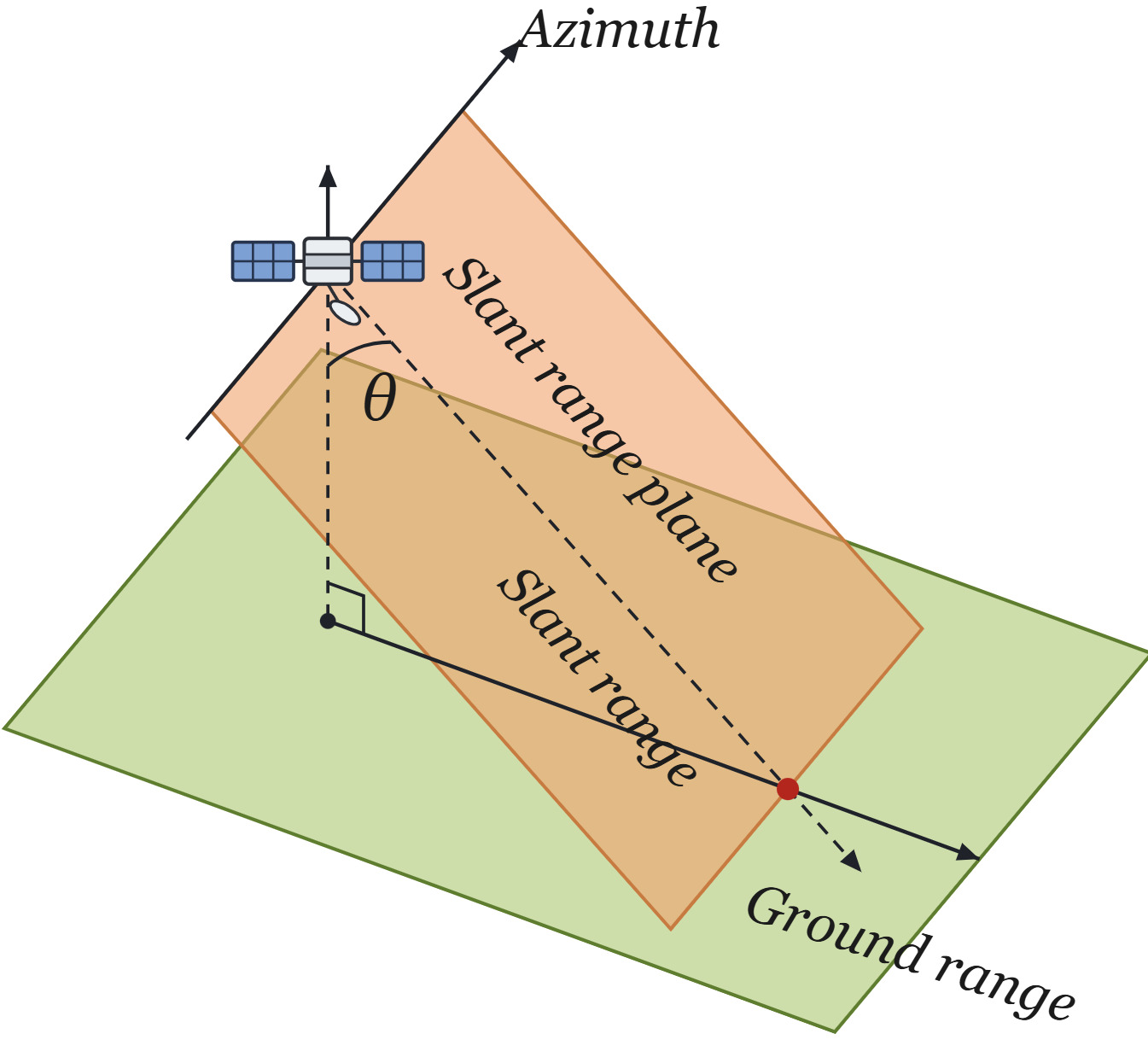}
    \caption{SAR geometry acquisition with slant-range and ground-range planes.}
    \label{fig:sar_opt_geometry}
  \end{subfigure}
  \caption{Overview of dataset coverage and SAR acquisition geometry.}
  \label{fig:umbra_and_geometry}
\end{figure}

We built our dataset from open-access spotlight acquisitions collected by the Umbra satellite constellation and distributed as \emph{Sensor Independent Complex Data} (SICD). SICD products provide complex-valued SAR images (magnitude and phase) together with rich metadata (sampling spacings, scene center point, imaging geometry), which makes them well suited for reproducible preprocessing. We select 2{,}565 SICD scenes covering all continents and diverse environments (urban, rural, coastal, mountainous) as shown in Figure~\cref{fig:umbra_locations}, with VV or HH polarization, incidence angles ranging from $10^\circ$ to $70^\circ$, and native resolutions from 20 cm to 2 m.

\begin{figure*}[t]
\centering

\begin{subfigure}[t]{0.49\textwidth}
  \centering
  \adjustbox{max width=\linewidth}{
    \includegraphics{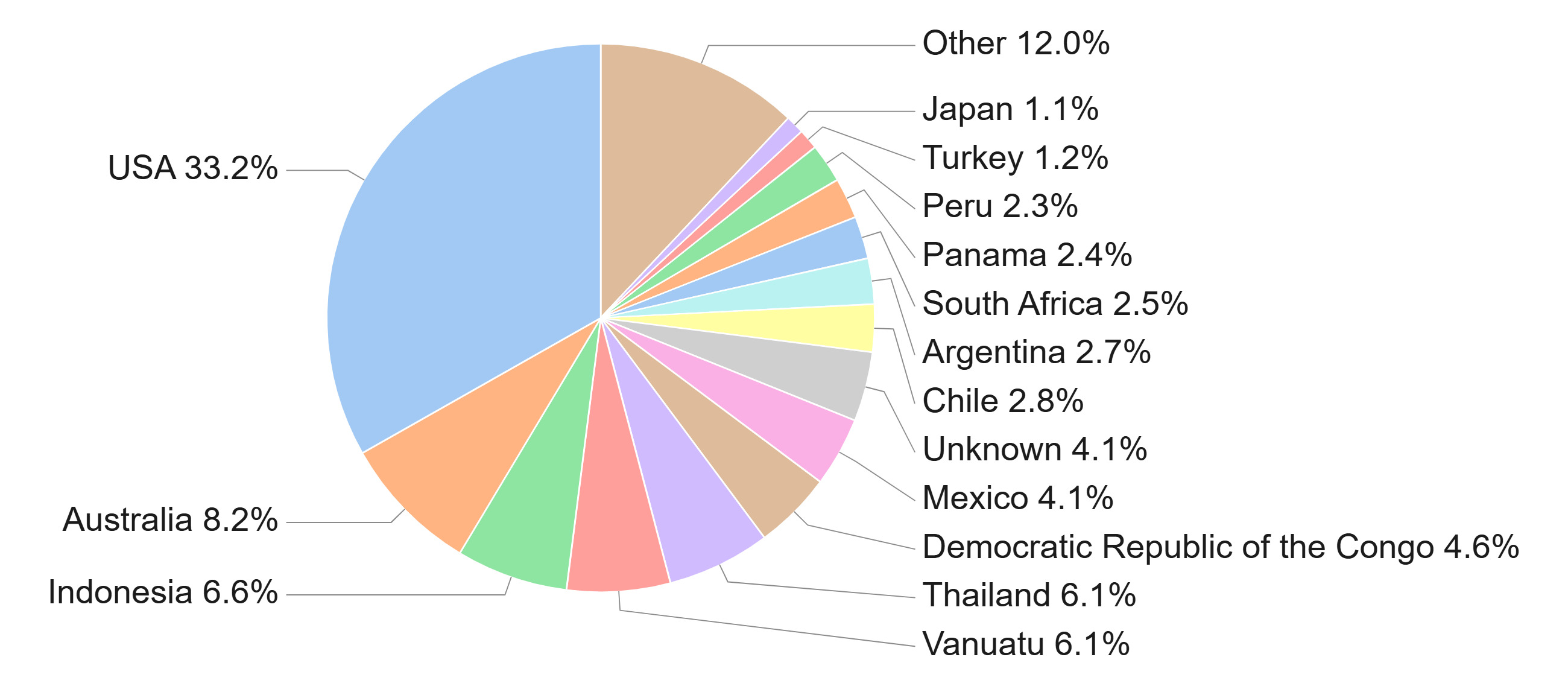}
  }
  \caption{Number of samples per country}
  \label{fig:imagettes_country}
\end{subfigure}
\hspace{1.5pt}
\begin{subfigure}[t]{0.49\textwidth}
  \centering
  \adjustbox{max width=\linewidth}{
    \includegraphics{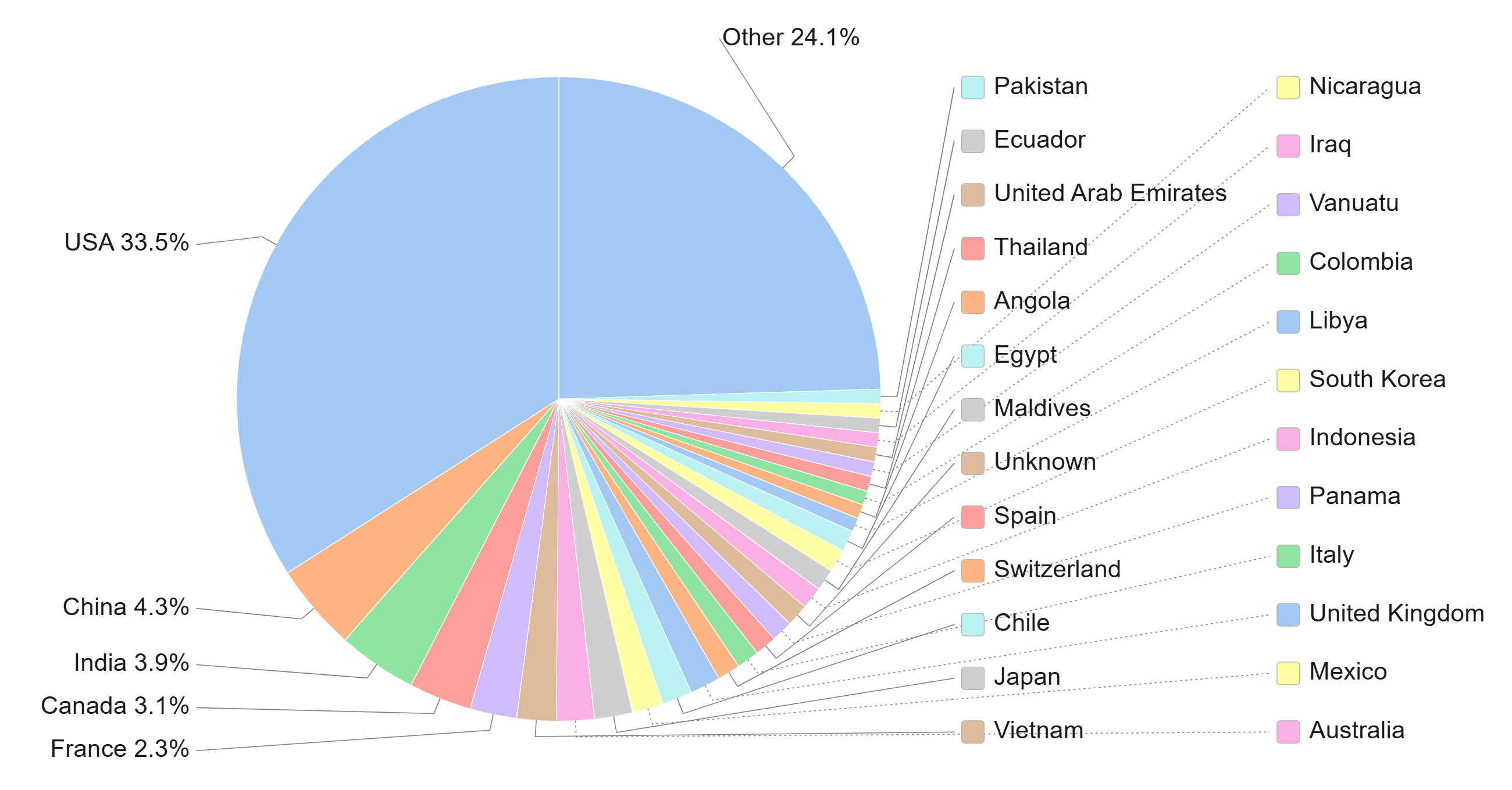}
  }
  \caption{Distinct cities per country}
  \label{fig:cities_country}
\end{subfigure}

\caption{Overview of our final dataset (Top-N + Other): distribution of images by country (left), and distribution of distinct cities by country (right)}
\label{fig:dataset_overview}
\end{figure*}

\begin{figure*}[t]
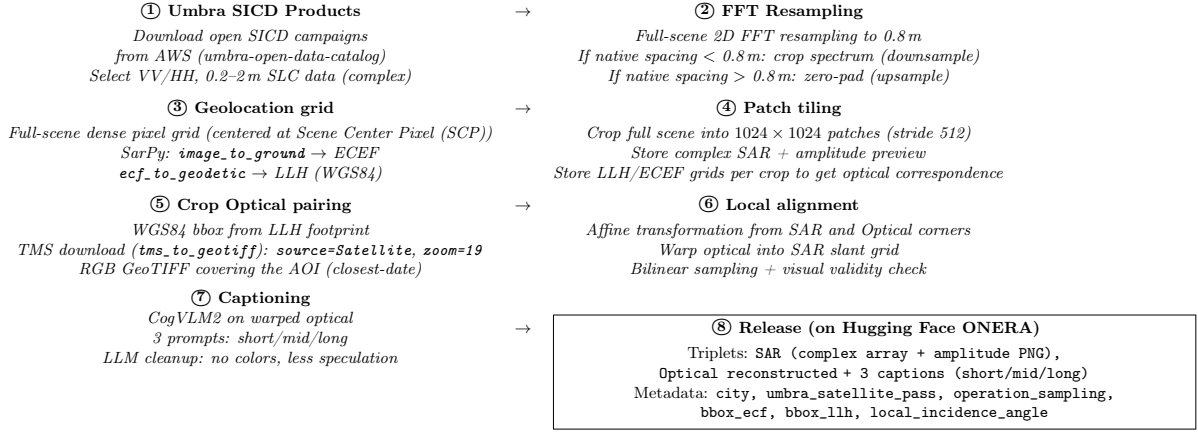

\centering
\footnotesize
\setlength{\tabcolsep}{6pt}
\renewcommand{\arraystretch}{1.25}

\begin{adjustbox}{max width=\textwidth}
\begin{tabular}{c c c c}

\textbf{\circnum{1} Umbra SICD Products} &
$\rightarrow$ &
\textbf{\circnum{2} FFT Resampling} & \\[2pt]

\makecell[c]{\textit{Download open SICD campaigns}\\
\textit{from AWS (umbra-open-data-catalog)}\\
\textit{Select VV/HH, 0.2--2\,m SLC data (complex)}} &
{} &
\makecell[c]{\textit{Full-scene 2D FFT resampling to 0.8\,m}\\
\textit{If native spacing $<$ 0.8\,m: crop spectrum (downsample)}\\
\textit{If native spacing $>$ 0.8\,m: zero-pad (upsample)}} &
{} \\

\noalign{\vskip 5pt}

\textbf{\circnum{3} Geolocation grid} &
$\rightarrow$ &
\textbf{\circnum{4} Patch tiling} & \\[2pt]

\makecell[c]{\textit{Full-scene dense pixel grid (centered at Scene Center Pixel (SCP))}\\
\textit{SarPy: \texttt{image\_to\_ground} $\rightarrow$ ECEF}\\
\textit{\texttt{ecf\_to\_geodetic} $\rightarrow$ LLH (WGS84)}} &
{} &
\makecell[c]{\textit{Crop full scene into $1024\times1024$ patches (stride 512)}\\
\textit{Store complex SAR + amplitude preview}\\
\textit{Store LLH/ECEF grids per crop to get optical correspondence}} &
{} \\

\noalign{\vskip 5pt}

\textbf{\circnum{5} Crop Optical pairing} &
$\rightarrow$ &
\textbf{\circnum{6} Local alignment} & \\[2pt]

\makecell[c]{\textit{WGS84 bbox from LLH footprint}\\
\textit{TMS download (\texttt{tms\_to\_geotiff}): \texttt{source=Satellite}, \texttt{zoom=19}}\\
\textit{RGB GeoTIFF covering the AOI (closest-date)}} &
{} &
\makecell[c]{\textit{Affine transformation from SAR and Optical corners}\\
\textit{Warp optical into SAR slant grid}\\
\textit{Bilinear sampling + visual validity check}} &
{} \\

\noalign{\vskip 5pt}

\makecell[c]{\textbf{\circnum{7} Captioning}\\
\textit{CogVLM2 on warped optical}\\
\textit{3 prompts: short/mid/long}\\
\textit{LLM cleanup: no colors, less speculation}} &
$\rightarrow$ &
\multicolumn{2}{c}{%
\fbox{\begin{minipage}[t]{0.77\textwidth}
\centering
\textbf{\circnum{8} Release (on Hugging Face ONERA)}\\
\smallskip
\makecell{Triplets: \texttt{SAR (complex array + amplitude PNG),}\\
{\texttt{Optical reconstructed}} \texttt{+ 3 captions (short/mid/long)}\\
Metadata: \texttt{city, umbra\_satellite\_pass, operation\_sampling,}\\
\texttt{bbox\_ecf, bbox\_llh, local\_incidence\_angle}}
\end{minipage}}%
} \\

\end{tabular}
\end{adjustbox}

\caption{Detailed overview of the dataset creation pipeline. Steps \circnum{1}--\circnum{7} describe downloading, preprocessing, pairing, and caption generation; step \circnum{8} summarizes released triplets.}
\label{fig:pipeline_overview}
\end{figure*}

As shown in Figure~\ref{fig:dataset_overview}, and reflecting Umbra’s acquisition footprint together with our processing steps (summarized in Figure ~\ref{fig:pipeline_overview}), the dataset has worldwide scope with 257 different locations in 72 countries. The USA contributes $\sim$33\% of samples and a large share of distinct cities, while the remaining images come from many countries with a long-tail distribution.

\subsection{Preprocessing steps}

\paragraph{\textbf{Frequency-domain resampling to 0.8 m pixel spacing}} Each SICD scene is converted into a fixed target sampling of 80 cm $\times$ 80 cm in slant-range geometry. To resample consistently while preserving the complex signal structure, we operate in the 2D frequency domain. We compute the centered spectrum then perform \emph{band-limited resampling} by either cropping the central band (downsampling) or zero-padding the spectrum (upsampling), depending on the original sampling compared to the target. Finally, the standardized complex image is obtained by the inverse Fourier Transform. This procedure preserves the band-limited coherent field and ensures Shannon–Nyquist-consistent resampling without introducing geometric distortions. 

\paragraph{\textbf{Geolocation grid: from SAR pixels to Earth coordinates}} SAR images are delivered in slant-range geometry, meaning that each pixel is indexed in the radar acquisition coordinates (azimuth and range) rather than in a map-projected ground plane, as shown in Figure \cref{fig:sar_opt_geometry}. 

To link each standardized SAR pixel to its physical location on Earth, we compute a dense geolocation grid using the SICD metadata. For each scene, the SICD file provides the scene center point (SCP) in pixel coordinates, the sampling spacings in azimuth and range, and the imaging geometry needed to project image coordinates onto the Earth. We first build a regular grid of pixel coordinates over the standardized SAR image, centered on the SCP and expressed in the original SICD sampling. We then apply the SICD image-to-ground projection to every pixel to obtain its corresponding Earth-Centered, Earth-Fixed (ECEF) coordinate. Finally, we convert these ECEF coordinates to WGS84 geodetic coordinates (latitude, longitude, altitude). This produces two aligned grids per scene (ECEF and lat/lon/height), which are saved once and later cropped to match each $1024 \times 1024$ SAR patch.

\paragraph{\textbf{Patch selection and tiling}} Each standardized scene is divided into overlapping patches of size $1024 \times 1024$ with a stride of 512 pixels. For each selected crop, we store the complex SAR patch, a normalized amplitude PNG, and the corresponding LLH/ECEF coordinate crops. 

\paragraph{\textbf{Optical pairing and projection into SAR slant-range geometry}} The dataset is SAR-oriented, so our goal is to keep SAR SLC data in its native slant-range geometry. Optical images are only approximately projected onto the SAR grid via coordinate correspondences (affine mapping).

The dataset is SAR-oriented: we keep SAR SLC patches in their native slant-range grid to preserve radar physics and avoid any resampling that could alter complex statistics. To build multimodal pairs, we associate each SAR patch with an optical RGB tile covering the same geographic area, then warp the optical image onto the SAR pixel grid using a simple coordinate-based approximation.



For each  1024×1024 SAR crop, we compute its geographic footprint from the LLH grid, retrieve the corresponding optical tile in WGS84, and estimate a first-order 2D affine transform from the SAR crop corners mapped into optical pixel coordinates. This affine model is a local approximation (valid over the small crop extent) and does not require an explicit sensor model. We then apply inverse mapping and bilinear interpolation to sample the optical intensities on the SAR grid:
Let $\tilde{\mathbf{p}}_{\mathrm{sar}}=[x\;y\;1]^{\top}$ and
$\tilde{\mathbf{p}}_{\mathrm{opt}}=[u\;v\;1]^{\top}$ denote homogeneous pixel coordinates
in the SAR and optical images, respectively.
From three (or four) corner correspondences
$\{(\mathbf{p}_{\mathrm{sar}}^{(i)},\mathbf{p}_{\mathrm{opt}}^{(i)})\}_{i=1}^{N}$, we estimate
the affine transform from SAR to optical,
\begin{equation}
\tilde{\mathbf{p}}_{\mathrm{opt}} \;=\; \mathbf{A}_{\mathrm{sar}\rightarrow\mathrm{opt}}\,\tilde{\mathbf{p}}_{\mathrm{sar}},
\qquad
\mathbf{A}_{\mathrm{sar}\rightarrow\mathrm{opt}}=
\begin{bmatrix}
a_{11} & a_{12} & t_x\\
a_{21} & a_{22} & t_y\\
0      & 0      & 1
\end{bmatrix},
\end{equation}
by solving a (least-squares) fit over the matched points.

To obtain an optical image on the SAR pixel grid, we use inverse warping
(i.e., we iterate over SAR output pixels and sample the optical source image).
For each SAR pixel $(x,y)$, we compute its corresponding optical coordinates
\begin{equation}
\begin{bmatrix}
u(x,y)\\ v(x,y)\\ 1
\end{bmatrix}
=
\mathbf{A}_{\mathrm{sar}\rightarrow\mathrm{opt}}
\begin{bmatrix}
x\\ y\\ 1
\end{bmatrix},
\end{equation}
and define the warped optical image as
\begin{equation}
I_{\mathrm{opt}\rightarrow\mathrm{sar}}(x,y)
=
I_{\mathrm{opt}}\!\big(u(x,y),\,v(x,y)\big),
\end{equation}
where $I_{\mathrm{opt}}$ is evaluated at non-integer coordinates using bilinear interpolation.
Pixels mapped outside the optical tile are set to zero.

\subsection{SAR--Optical pairs: Captioning}
\begin{figure}[ht]
  \centering
  \includegraphics[width=\columnwidth,trim=20 10 20 10,clip]{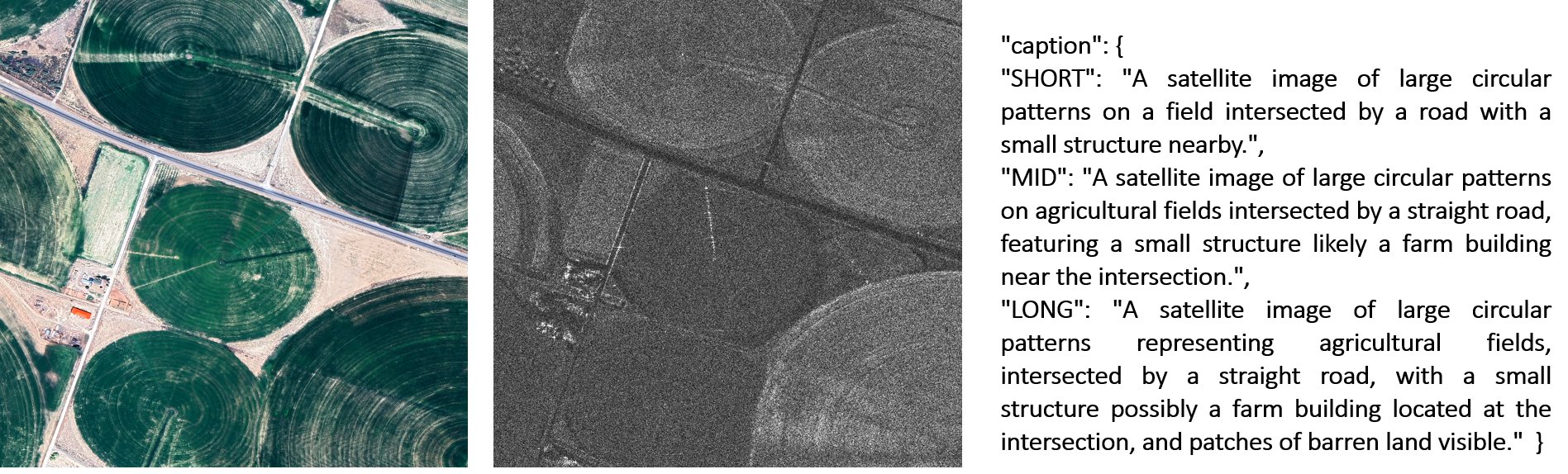}
  \caption{Example of SAR--optical--Caption from our Dataset}
  \label{fig:3_captions_SAR_Opt_Text}
\end{figure}

Captions are generated directly from the optical patch projected to the SAR grid. We use CogVLM2 with three prompting variants to obtain three captions per patch (short, medium, and long). All prompts follow three different instructions, for example: \emph{`Describe the key structural features of the satellite image in a few words. Do not use color terms in the text description. Avoid hypothesis or vague wording.''}. We then post-process and normalize the generated captions using a large language model to improve consistency, remove residual color words, and reduce speculative phrasing. Figure~\cref{fig:3_captions_SAR_Opt_Text} illustrates one SAR-optical pair along with example captions such as: \emph{A satellite image of large circular patterns of a field intersected by a road with a small structure nearby.}

\begin{figure*}[t]
\centering

\begin{subfigure}[t]{0.55\textwidth}
  \vspace{0pt}
  \centering
  \adjustbox{max width=\linewidth}{
    \includegraphics{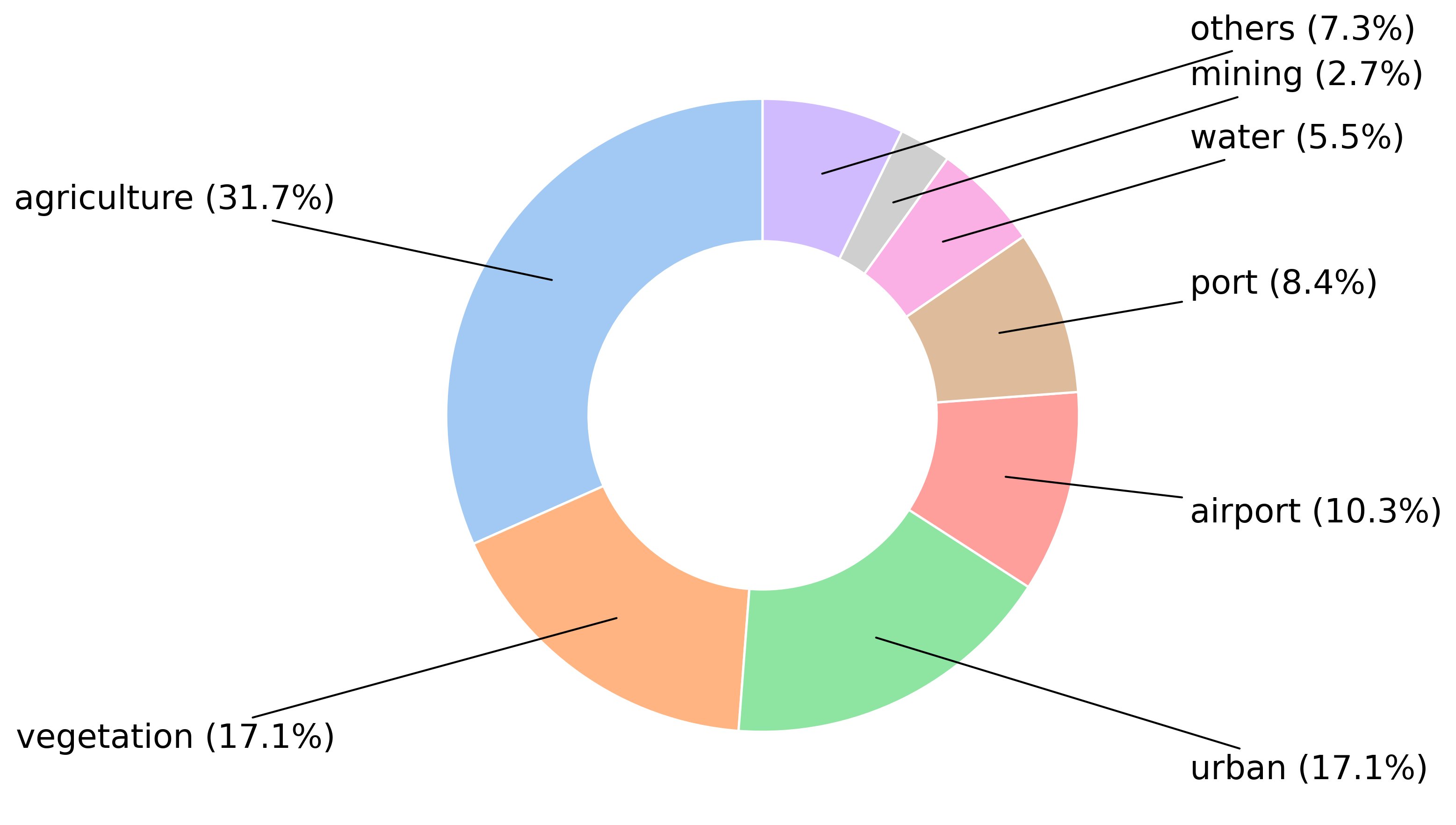}
  }
  \caption{Label distribution from prompt-based keyword matching.}
  \label{fig:captions_labels}
\end{subfigure}
\hfill
\begin{subfigure}[t]{0.40\textwidth}
  \vspace{0pt}
  \centering
  \small
  \adjustbox{max width=\linewidth}{
  \begin{tabular}{lrrrrr}
    \toprule
    \textbf{Prompt} & \textbf{Avg.} & \textbf{P10} & \textbf{P90} & \textbf{Vocab} & \textbf{Dup.} \\
    \midrule
    SHORT & 14.9 & 12 & 18 & 1623 & 0.2643 \\
    MID   & 25.3 & 20 & 32 & 3376 & 0.0044 \\
    LONG  & 37.1 & 27 & 49 & 4655 & 0.0000 \\
    \bottomrule
  \end{tabular}}
  \caption{Prompt text statistics: Avg.\ words per prompt; P10/P90 are the 10th/90th percentiles of prompt length (in words); Vocab is the vocabulary size; Dup.\ rate is the fraction of prompts that are exact duplicates.}
  \label{fig:prompt_stats_side}
\end{subfigure}

\caption{Prompt labeling overview (left) and summary prompt statistics (right).}
\label{fig:labels_and_stats}
\end{figure*}

Across 119,566 patches, the three caption variants average 14.9/25.3/37.1 words (SHORT/ MID/ LONG) and show increasing lexical diversity, with a larger and more varied vocabulary in the longer captions (see Figure \ref{fig:prompt_stats_side}. Several qualitative examples from the dataset are shown in Fig.~\ref{fig:gallery_selected_pairs}.

We labeled each MID caption with a scene category using a manually defined keyword vocabulary (positive/negative terms) and a simple rule-based matcher, resulting in a diverse distribution of scene types, including agriculture, vegetation, urban areas, airports, ports, water, mining, volcanoes, and other landscapes like coastal, roads, volcano, industrial, mountain, desert, snow ice and rail (see Figure \ref{fig:prompt_stats_side}).

Thus, our dataset consists of 119,566 samples: a complex SAR patch (SICD-derived), its aligned optical image in slant-range geometry, and text descriptions. 

\section{Dataset Storage, Format, and Licensing}
\label{sec:storage}

\subsection{Storage and access}

We release SARLO-80 on the Hugging Face Hub as a WebDataset made of sharded
\texttt{.tar} archives. The dataset is about 1.42\,TB in total with optical reconstructed pairs. The files are
organized as \texttt{train/chunk\_{ID}/ shard-{ID}.tar} and are split into
train, validation, and test sets. The splits are disjoint and are made by
satellite pass. This format makes the dataset easier to use at large scale. Samples can be
streamed and shuffled directly from the archives, without extracting the full
dataset. Each sample has a unique key and contains the SAR data, the SICD
metadata, and the text annotations, as shown in Table~\ref{tab:sample_contents}.

The optical image is not stored as a warped PNG. Instead, we store the metadata needed to reconstruct it when needed. This keeps the dataset lighter and also avoids redistributing optical imagery from a third-party provider. The main fields used for this reconstruction are listed in Table~\ref{tab:sample_contents}. To reconstruct the optical view, the WGS84 footprint in \texttt{meta.json} is used to download the optical tile. Then, the four corners of the SAR crop are projected to WGS84 using the SICD metadata. These points are converted into pixels in the optical image. An affine transform is then estimated and used to warp the optical tile into the SAR crop frame. Since this transform is stored in the metadata, users can also project other external maps or labels into the SAR grid. We provide reference code for this reconstruction.

\begin{table}[t]
\centering
\small
\setlength{\tabcolsep}{5pt}
\renewcommand{\arraystretch}{1.2}
\begin{tabular}{l p{0.62\linewidth}}
\toprule
\textbf{File} & \textbf{Content} \\
\midrule
\texttt{<id>.sar.jpg} & SAR amplitude image in slant-range geometry ($\sim$1024$\times$1024) \\
\texttt{<id>.sar.npy} & Complex-valued SAR array in slant-range geometry \\
\texttt{<id>.sicd.xml} & SICD metadata of the original Umbra acquisition \\
\texttt{<id>.meta.json} & Geometry, captions, incidence angles, and optical reconstruction metadata: the WGS84 corners, source, and zoom level of the optical tile; the SAR crop indices, size and sampling (downsampling or upsampling) ; the scene-center pixel coordinates; the crop bounding boxes in ECEF and latitude/longitude/height; the incidence angles (terrain, ellipsoid, and SICD, when available); and three caption versions (short, mid, long) \\
\texttt{<id>.\_\_key\_\_} & Unique WebDataset sample key \\
\bottomrule
\end{tabular}
\caption{Contents of each released WebDataset sample. The optical PNG is not redistributed. It is reconstructed when needed from \texttt{meta.json} and \texttt{sicd.xml}.}
\label{tab:sample_contents}
\end{table}

\subsection{Licensing}

SARLO-80 combines data from different sources. The SAR data come from the Umbra Open Data Program. The original SICD
products are released under a Creative Commons Attribution license
(CC-BY-4.0), with attribution to Umbra. We therefore redistribute the resampled SAR patches and the related SICD metadata with this attribution. The optical images are retrieved from Google Satellite TMS at zoom level 19. For this reason, we only provide the metadata and reference code needed to download the optical tiles and project them into the SAR grid. 

\section{Experiments and Discussion}
\label{sec:experiments}
Recent progress in vision--language models (VLMs) and generative text-to-image pipelines has demonstrated that large-scale multimodal alignment enables strong cross-modal retrieval, captioning, and conditional generation capabilities. However, these advances have been almost exclusively developed for optical imagery. In contrast, very-high-resolution (VHR) synthetic aperture radar (SAR) imagery, particularly in complex-valued SLC format, remains largely unexplored in multimodal learning benchmarks.

Existing SAR datasets typically rely on lower-resolution GRD products or different sensor configurations, making direct comparison with current foundation models difficult. As a result, there are currently no publicly available vision--language models trained on VHR SAR SLC imagery. To illustrate the capabilities enabled by the dataset, we evaluate two representative multimodal tasks: text-to-SAR generation.

\subsection{Text-to-SAR Generation Baseline}
In previous work \cite{DEBUYSERE202693}, we adapt a text-to-image latent diffusion model \cite{podell2023sdxlimprovinglatentdiffusion} using standard parameter-efficient fine-tuning and prompt augmentation strategies to benefit from its strong text–image composition. 

\begin{figure*}[t]
\centering
\setlength{\tabcolsep}{2pt}
\renewcommand{\arraystretch}{1}
\setlength{\gtilew}{0.192\textwidth}

\newcommand{\gtile}[1]{%
  \includegraphics[width=\gtilew,height=\gtilew]{#1}%
}

\begin{tabular}{*{5}{@{}c@{\hspace{2pt}}}}
\gtile{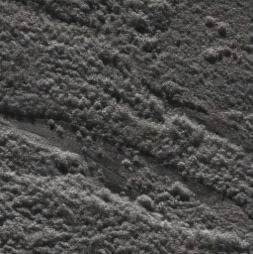} &
\gtile{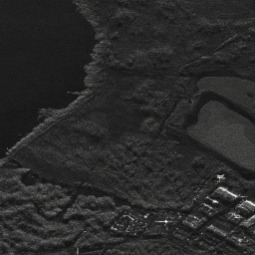} &
\gtile{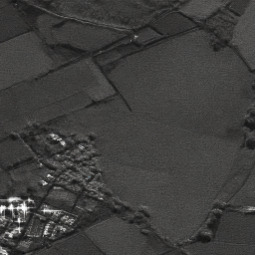} &
\gtile{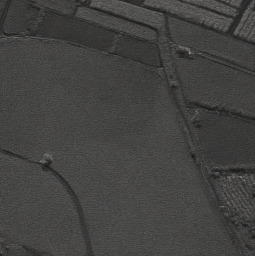} &
\gtile{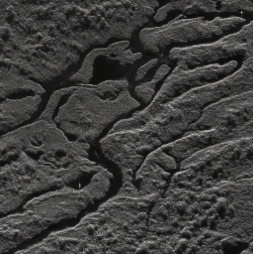} \\[2pt]

\gtile{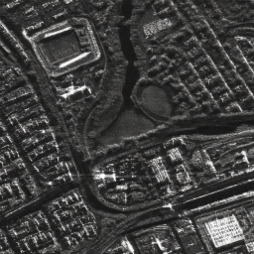} &
\gtile{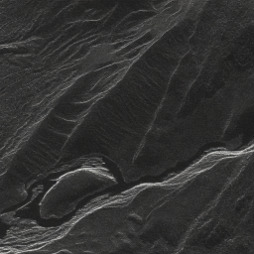} &
\gtile{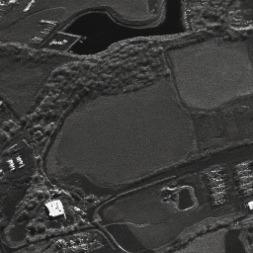} &
\gtile{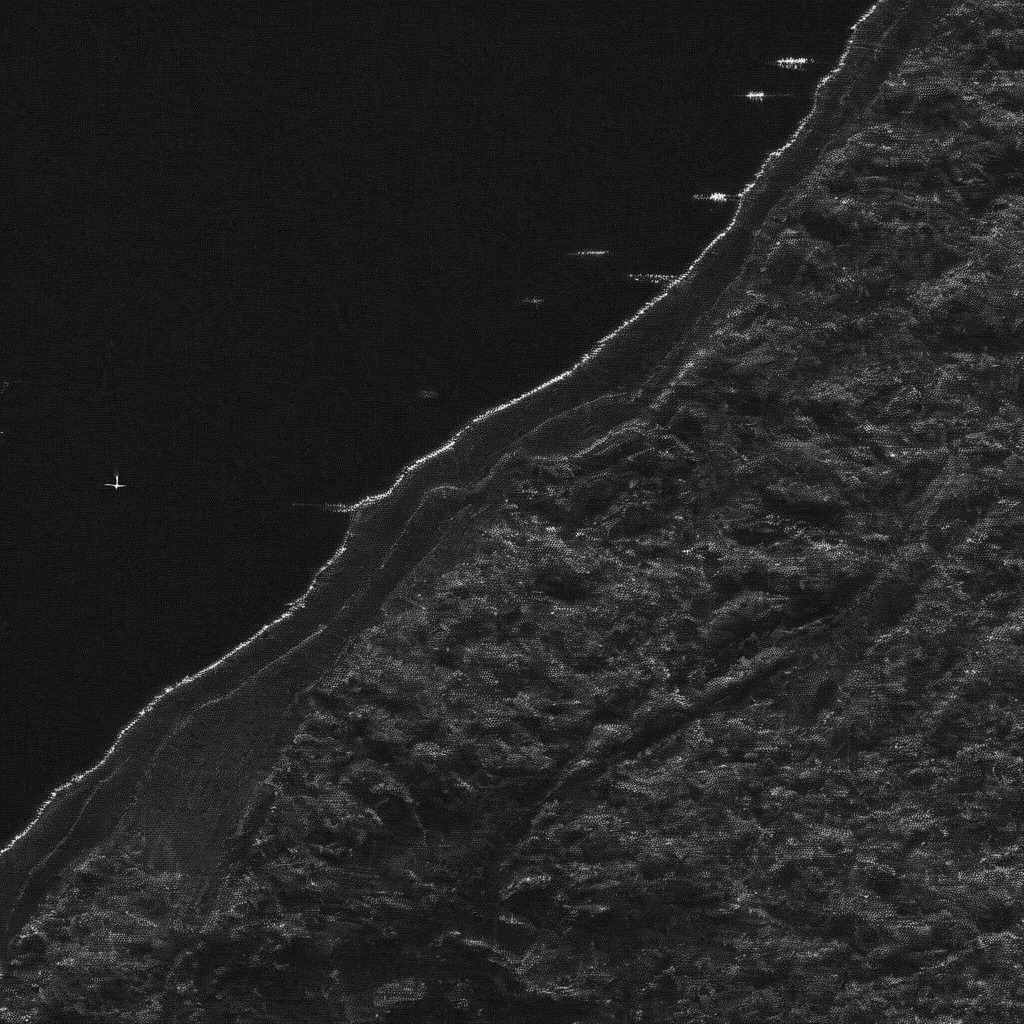} &
\gtile{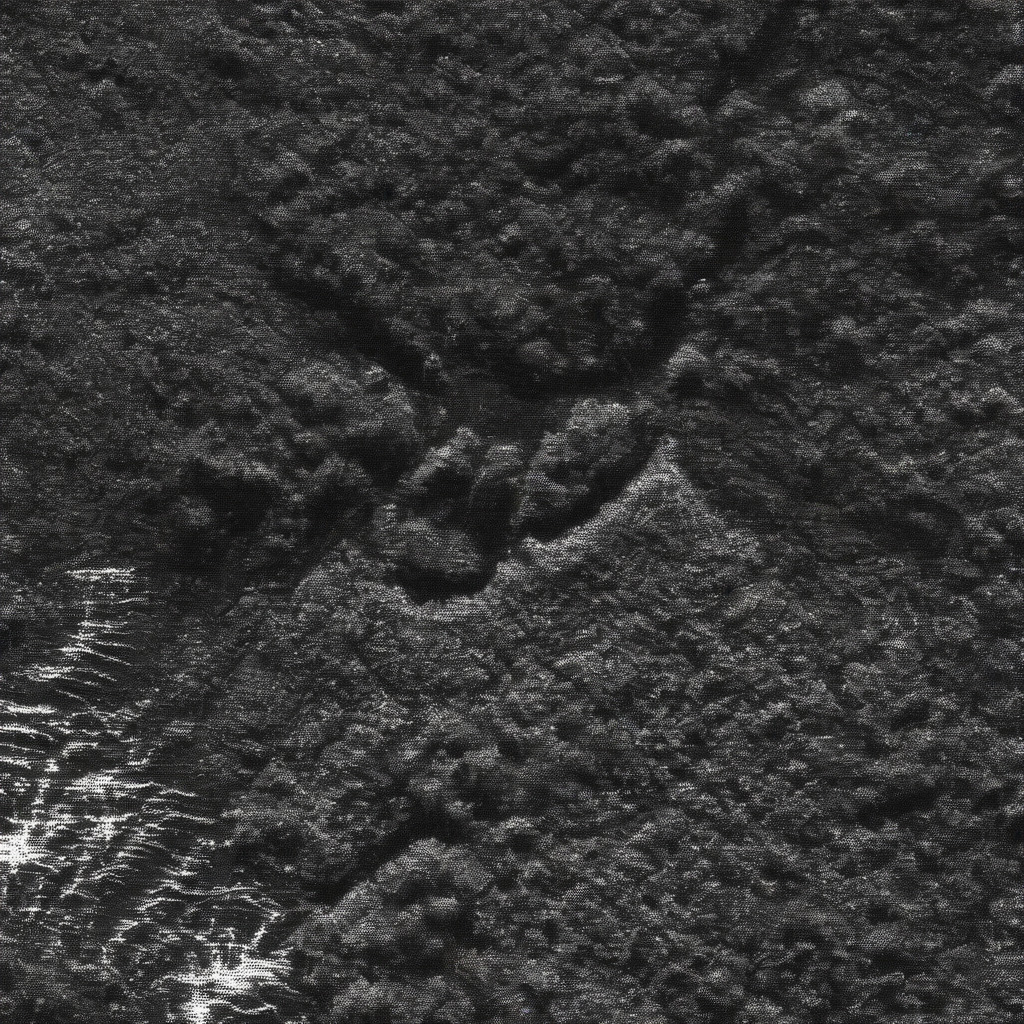} \\
\end{tabular}

\caption{Examples of generated SAR samples from fine-tuned SDXL backbone.}
\label{fig:generated_vrac}
\end{figure*}

\begin{figure*}[t]
\centering
\setlength{\tilew}{0.188\textwidth}

\newcommand{\tile}[1]{%
  \includegraphics[width=\tilew,height=\tilew]{#1}%
}

\newcommand{\bluetile}[1]{%
  {\setlength{\fboxsep}{0pt}%
   \setlength{\fboxrule}{2.5pt}%
   \fcolorbox{pipeEdge}{pipeFill}{%
     \includegraphics[
       width=\dimexpr\tilew-5pt\relax,
       height=\dimexpr\tilew-5pt\relax
     ]{#1}%
   }%
  }%
}

\newcommand{\bluearrow}{%
  \tikz{%
    \draw[
      -{Triangle[length=3mm,width=4.5mm]},
      pipeArrow,
      line width=3pt
    ] (0,0)--(0.74\textwidth,0);
  }%
}

\setlength{\tabcolsep}{3pt}
\renewcommand{\arraystretch}{1.1}

\begin{tabular}{*{5}{>{\centering\arraybackslash}m{\tilew}}}

\bluetile{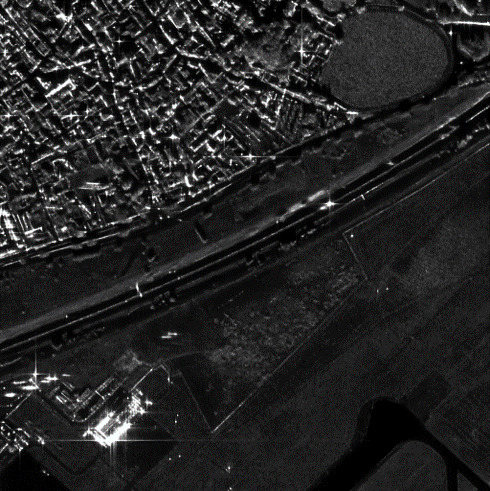} &
\tile{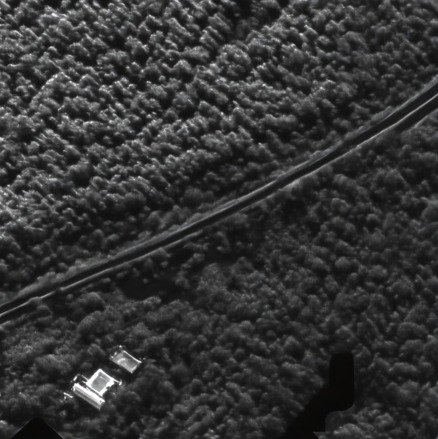} &
\tile{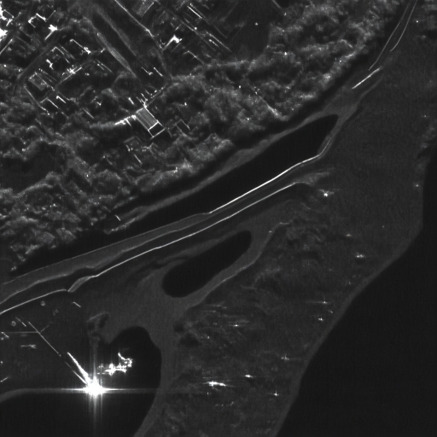} &
\tile{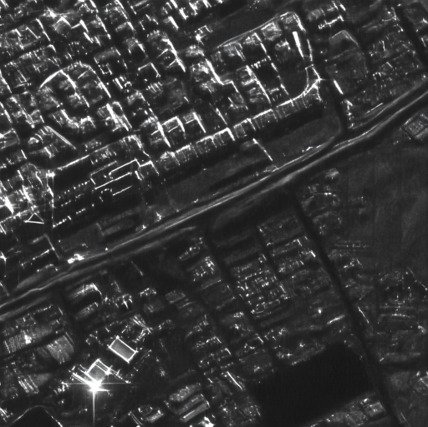} &
\tile{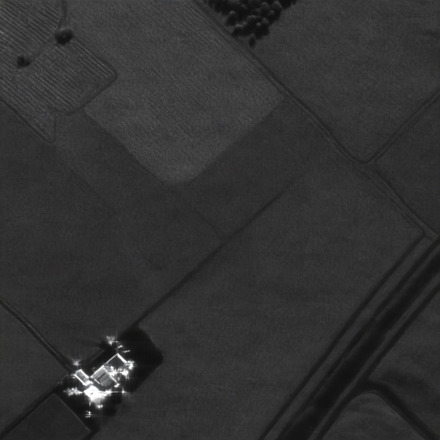} \\[2pt]

&
\multicolumn{4}{c}{\bluearrow} \\[2pt]

\bluetile{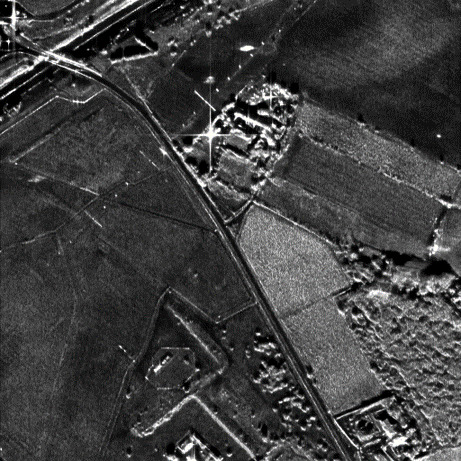} &
\tile{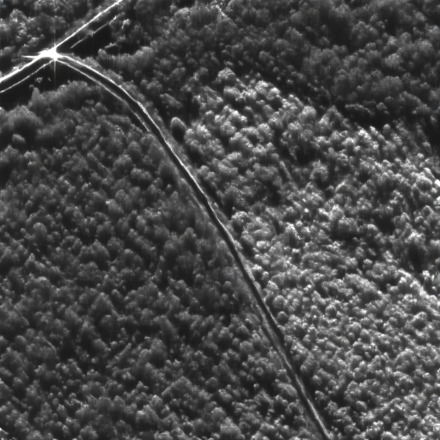} &
\tile{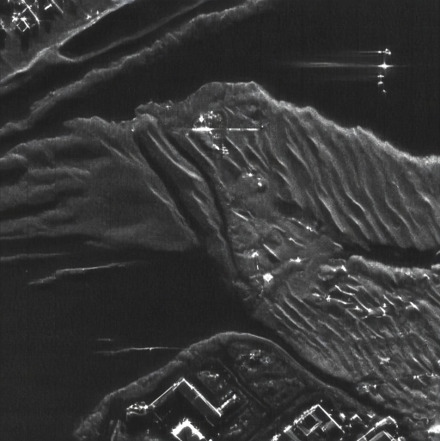} &
\tile{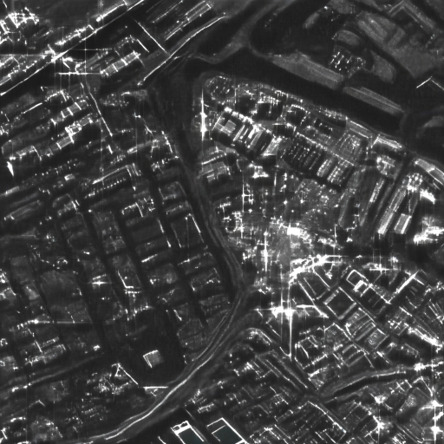} &
\tile{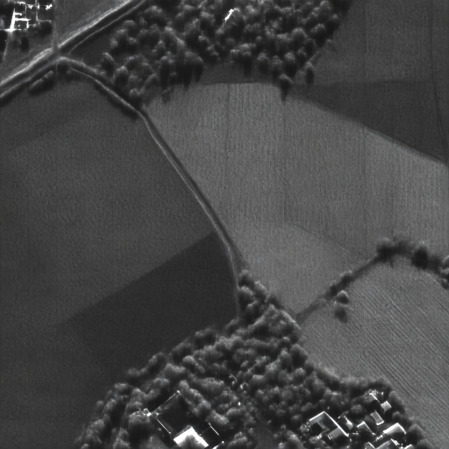} \\[2pt]

\scriptsize Real SAR SLC (80cm)&
\scriptsize a dense forest with a main road &
\scriptsize a coastal area with a boat &
\scriptsize a city with industrial buildings &
\scriptsize agricultural fields with a few isolated houses \\

\end{tabular}

\caption{Variants of real SAR SLC images generated with the fine-tuned SDXL backbone.
The two blue-framed tiles on the left show the real SAR references (\cite{sethi}); the remaining columns
show generated variants conditioned on the text below.}
\label{fig:sethi_variants}
\end{figure*}

In addition to direct generation from noise and textual description \ref{fig:generated_vrac}, this single generative backbone allows us to explore several spatial conditioning settings within the same framework. In particular, by conditioning the diffusion process on partial SAR observations, the model can be used to generate text-guided variants \ref{fig:sethi_variants}, as well as perform inpainting or outpainting as in the referenced method (citation in \cite{10993695} paper). This makes the approach especially relevant for data augmentation, since new samples can be produced either from text alone or from constrained spatial contexts while preserving control through the prompt. Moreover, the text-conditioning mechanism provides a natural way to analyze how semantic concepts are associated with SAR structures, for instance through the study of cross-attention maps.

\subsection{Discussion}

The results show that SARLO-80 can support meaningful alignment between SAR
images and natural language. At the same time, they also show that SAR remains a difficult modality to model, because its appearance is strongly linked to the physics of radar imaging. Some objects that are easy to describe in optical images can be much harder to recognize in SAR images, especially when their radar response is affected by geometry, speckle, or strong scattering effects. The 80\,cm resolution is also important. It is much finer than typical Sentinel-scale GRD products, so SAR-specific effects become more visible. These include layover, shadowing, strong urban scattering, and non-Gaussian scattering statistics, which can be studied for example with log-cumulant analysis. The incidence-angle metadata can also help analyze how acquisition geometry and terrain affect the SAR appearance.

The SAR--text pairs make it possible to study vision--language models. 
In previous work, we used SARLO-80 to fine-tune a vision--language model (Qwen2.5-VL) to caption SAR images, together with an airborne very-high-resolution SAR source. The result showed that a single model can handle both sensors at once: the spaceborne 80\,cm data and the airborne 40\,cm data, staying robust on each. Standard captioning metrics (METEOR, ROUGE-L, BERTScore) are not enough, because they reward the form of the sentence more than its match to the image; even captions paired with the wrong image score almost as high, so we instead compared captions with an LLM-as-judge that looks at meaning. Finally, training only on captioning does not erase the model's other skills (counting, question answering, other languages, unseen sensors); the only cost is a mild attachment to the training prompt, which is easily fixed. This shows SARLO-80 helps connect SAR images and language, and that evaluation needs as much care as the data.


Another important point is that the SAR patches are complex-valued and kept in slant-range geometry. This makes it possible to test different SAR
representations, such as amplitude, log-intensity, phase-related information, or other derived features. Comparing these representations can help identify which parts of the SAR signal are most useful for multimodal alignment and retrieval. The complex SAR signal can be used even without the optical image or the text caption. It supports tasks such as despeckling, spectrum-domain super-resolution, and speckle or equivalent-number-of-looks analysis. It can also be decomposed into sub-apertures. These sub-apertures can be used to study scattering diversity and to create colorized SAR images, by assigning different sub-aperture views to different color channels. This SAR colorization can make some structures easier to interpret and can also provide another useful representation for learning models.

The SAR--optical pair adds another useful direction. The optical image gives a
more intuitive view of the scene, while the SAR image shows radar-specific
effects. When an optical-based description does not fully match the SAR
appearance, this difference can itself be informative. It can reveal effects such
as layover, shadowing, or multi-bounce scattering, and can help models learn the
gap between optical and radar geometry.

Finally, we keep the tiles large, with a size of $1024\times1024$ pixels,
because small patches are often not enough for foundation-model training.
However, this also makes the problem harder. At very high resolution, speckle has a mixed behaviour: it is mostly random in homogeneous areas, but it can become more deterministic around strong urban scatterers. This remains difficult for standard generative models, including diffusion models. For this reason, complex-domain very-high-resolution SAR generation is still an open research problem.

\section{Conclusion}
\label{sec:conclusion}
To the best of our knowledge, this is the first large-scale dataset providing such a large collection (>$110{,}000$) of co-registered SLC SAR-optical patches at sub-meter resolution while preserving the complex SAR measurements. Overall, this dataset provides both semantic and geographic signals for downstream analysis and benchmarking. We expect our dataset to support research on multimodal representation learning, retrieval, segmentation, and generative modeling for SAR, and to serve as a foundation for training and evaluating SAR-oriented foundation models. Future work includes extending the dataset with additional modalities (e.g., elevation data or co-registered multi-temporal scenes), improving caption diversity and grounding, and introducing standardized downstream benchmarks and evaluation protocols tailored to very-high-resolution slant-range SAR.

\begin{figure*}[t]
\centering
\captionsetup{font=footnotesize}

\setlength{\tabcolsep}{2pt}
\renewcommand{\arraystretch}{1.0}

\begin{adjustbox}{width=\textwidth}
\begin{tabular}{@{}cccccc@{}}
    \includegraphics[width=0.16\textwidth]{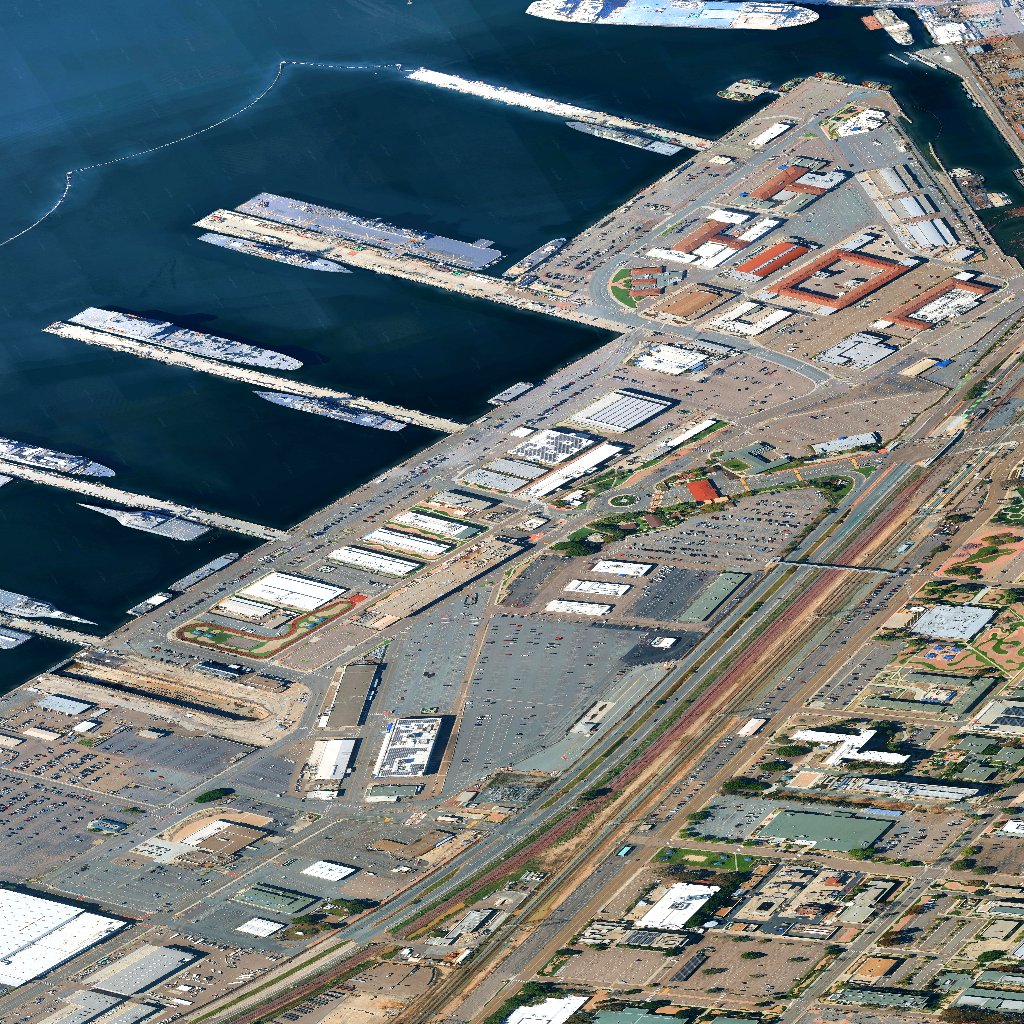}&\hspace{2pt}%
    \includegraphics[width=0.16\textwidth]{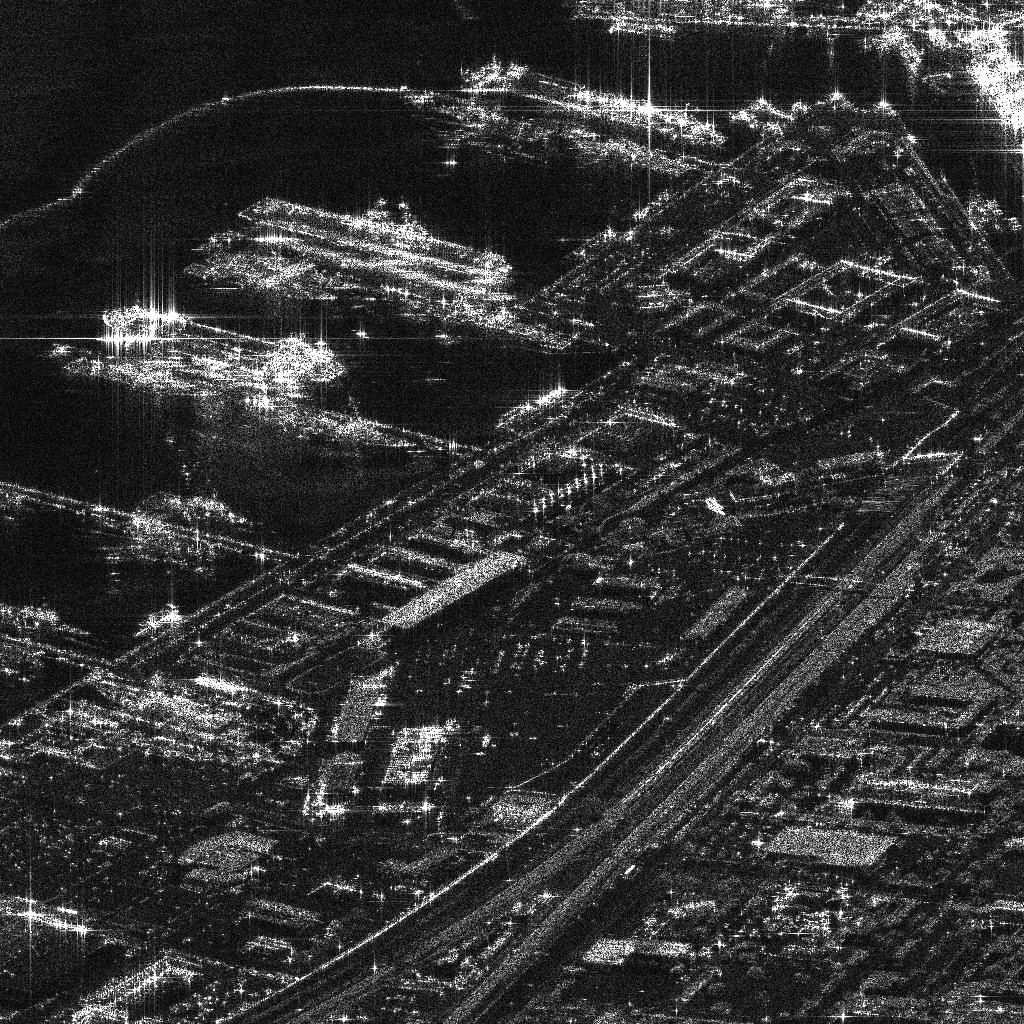}%
  &     \includegraphics[width=0.16\textwidth]{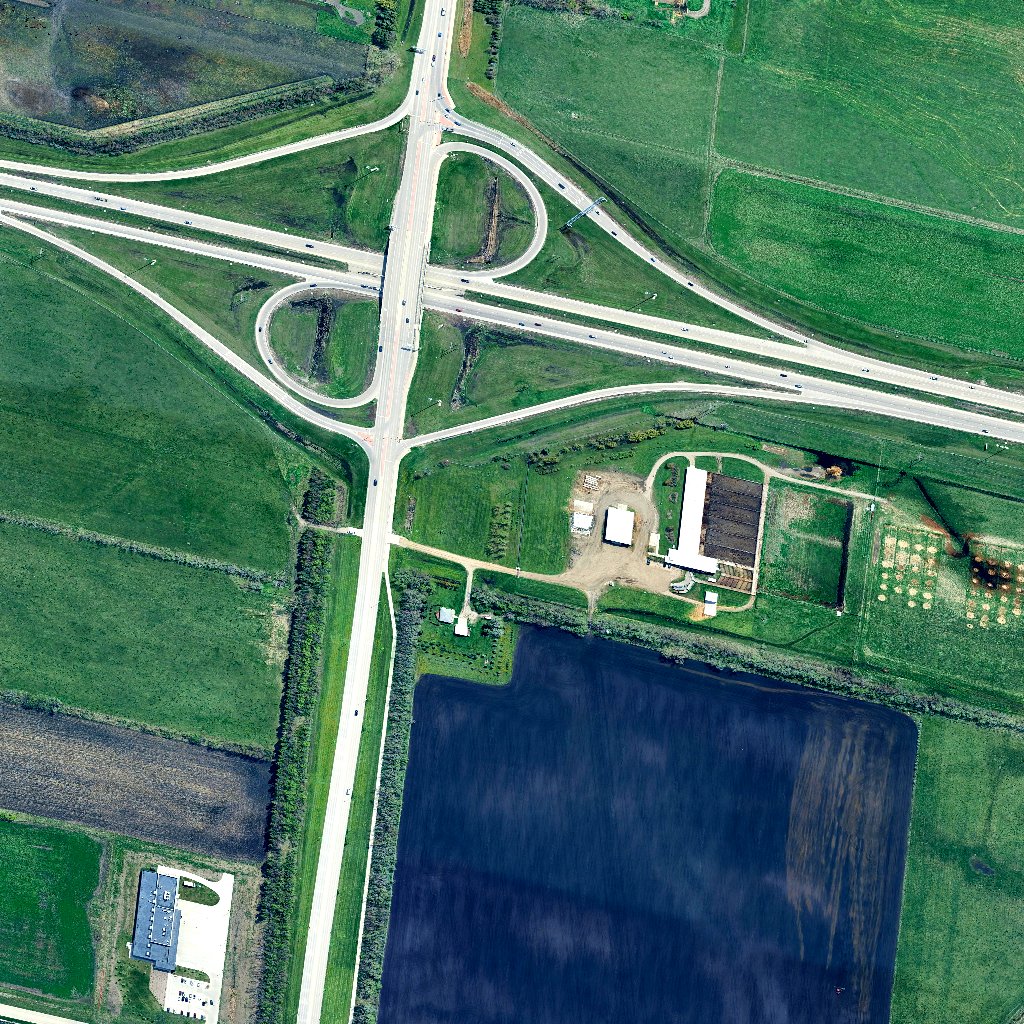}&\hspace{2pt}%
    \includegraphics[width=0.16\textwidth]{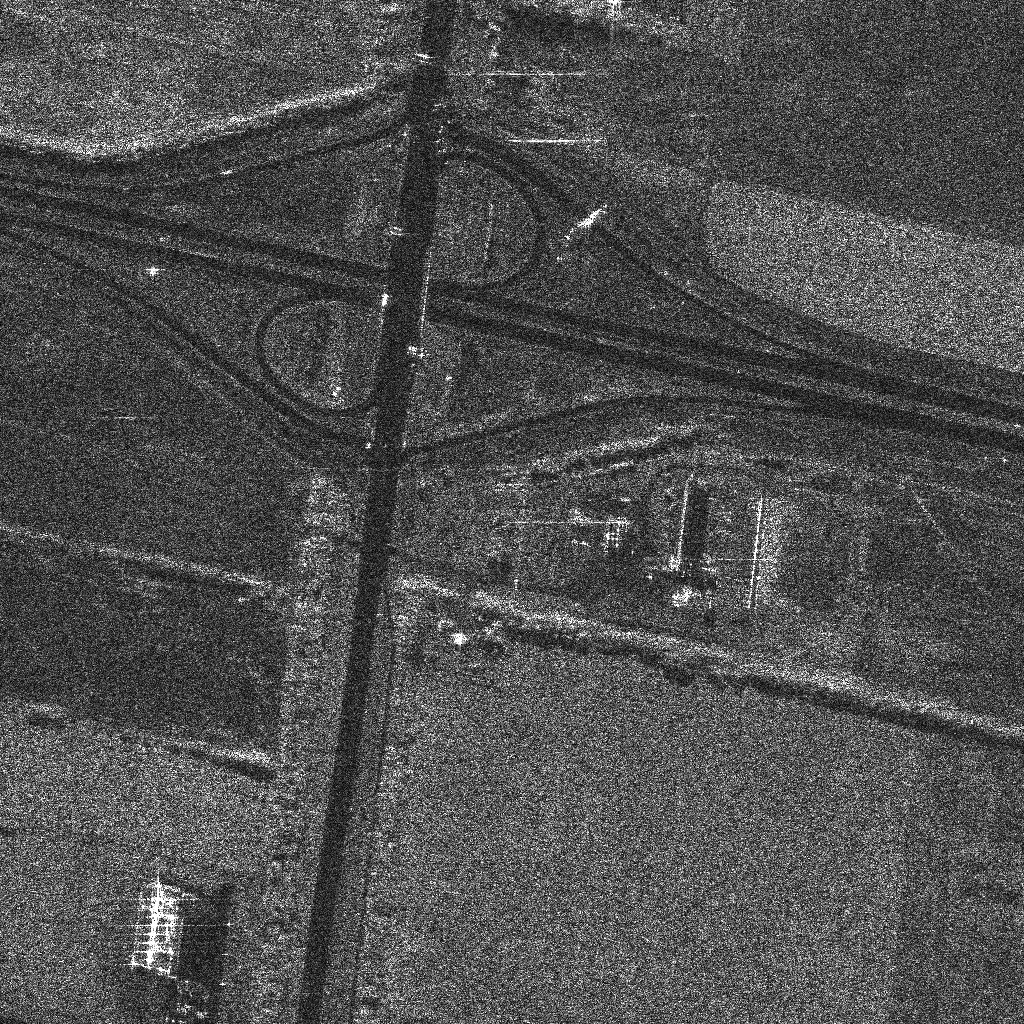}%
  &     \includegraphics[width=0.16\textwidth]{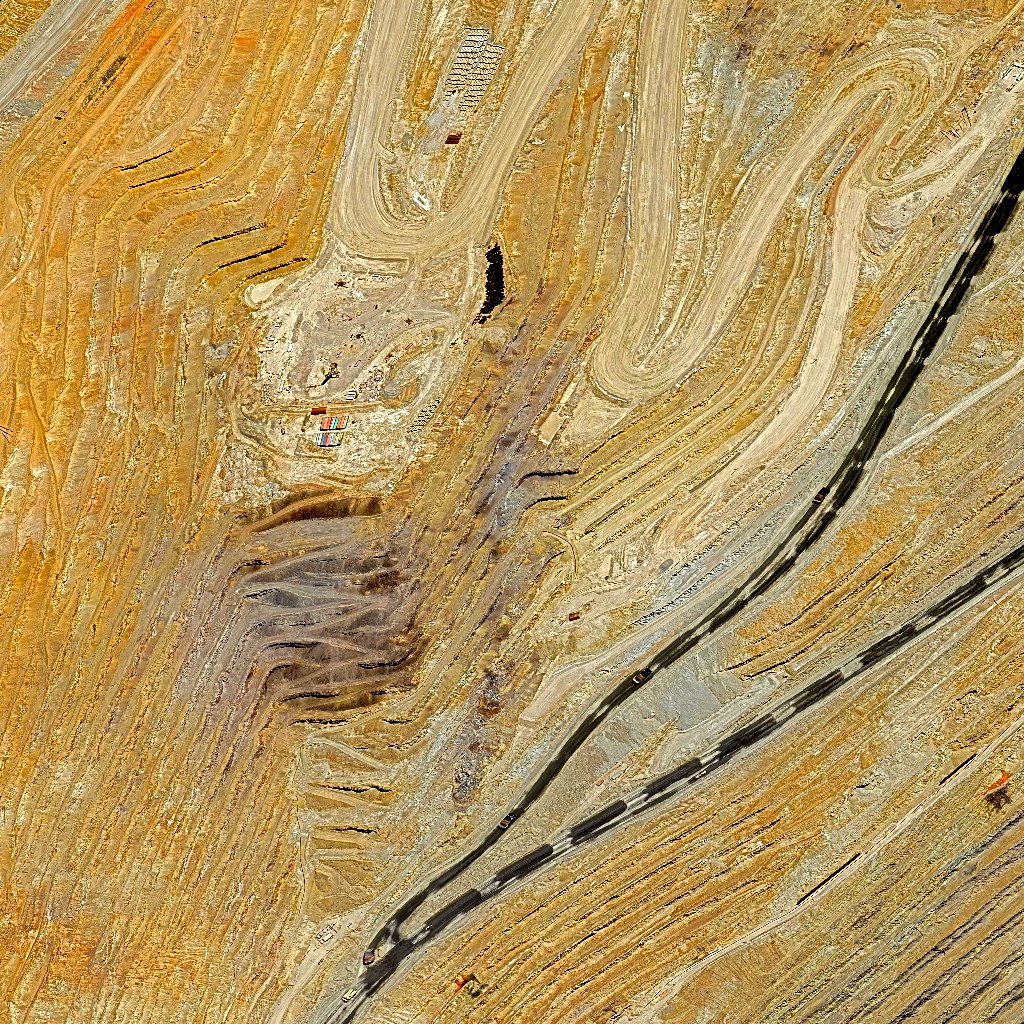}&\hspace{2pt}%
    \includegraphics[width=0.16\textwidth]{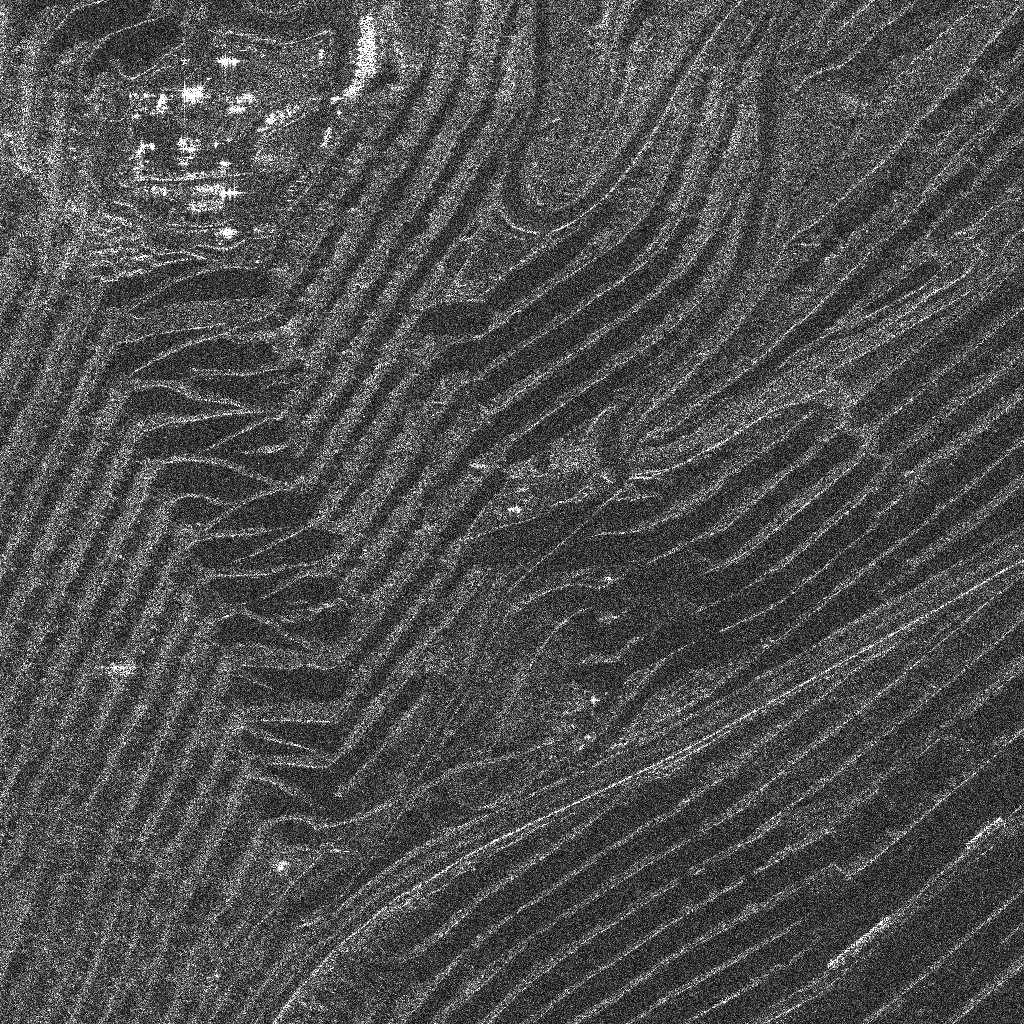}  \\
\multicolumn{2}{c}{\small (a)} & \multicolumn{2}{c}{\small (b)} & \multicolumn{2}{c}{\small (c)} \\[2pt]
    \includegraphics[width=0.16\textwidth]{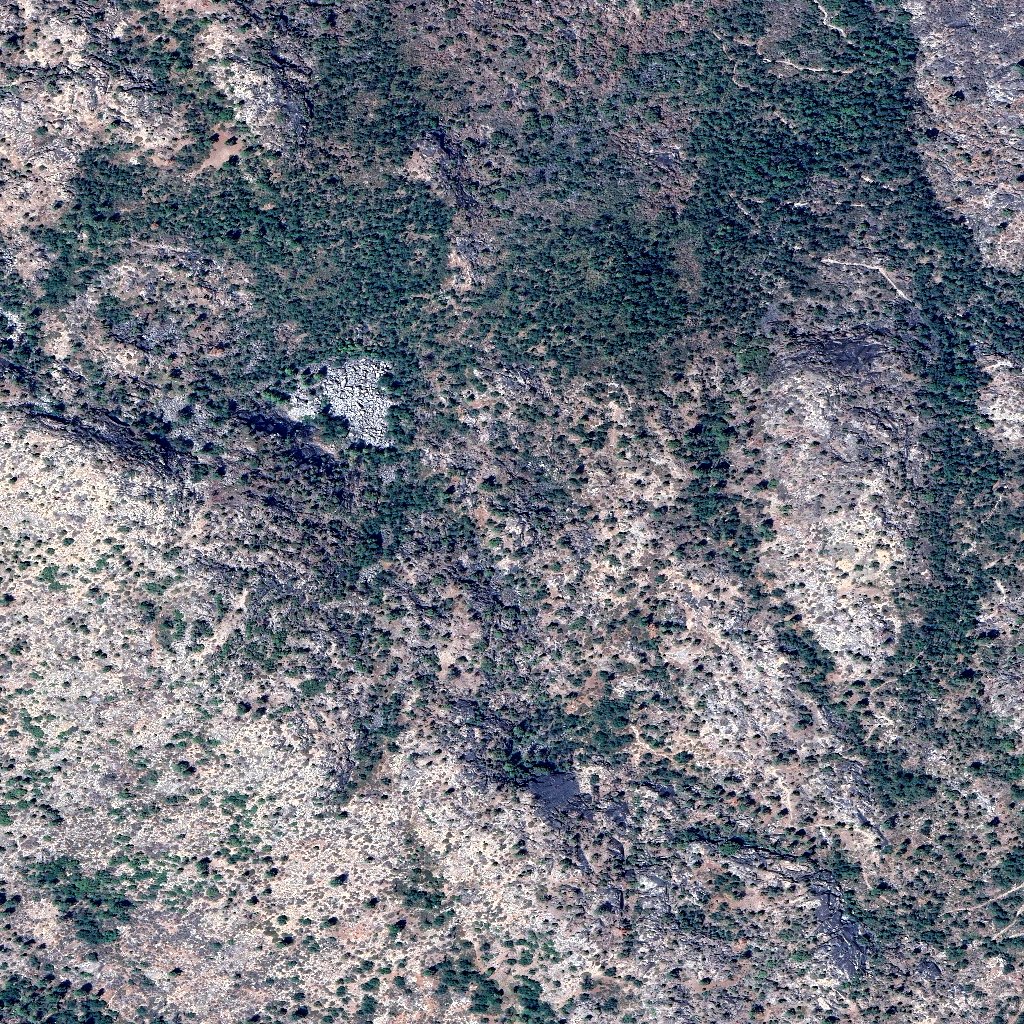}&\hspace{2pt}%
    \includegraphics[width=0.16\textwidth]{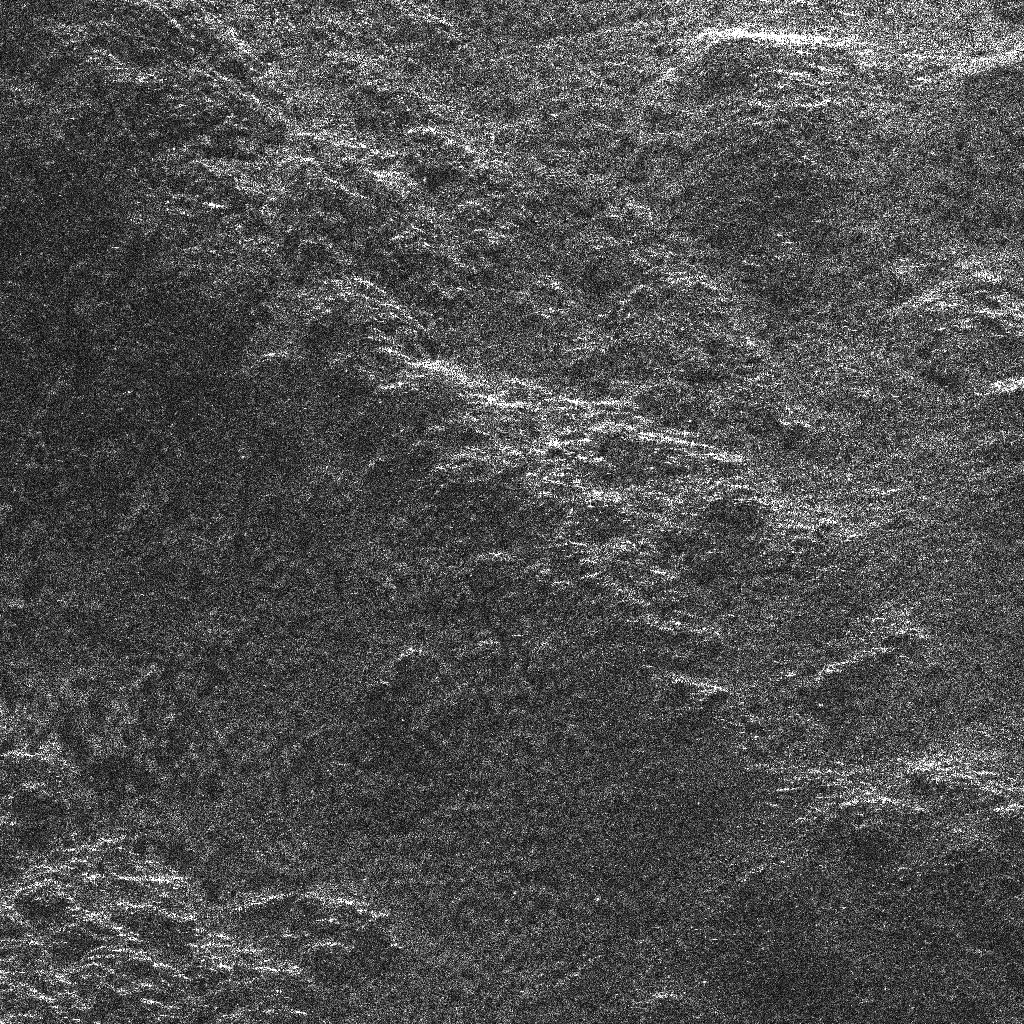}%
  &     \includegraphics[width=0.16\textwidth]{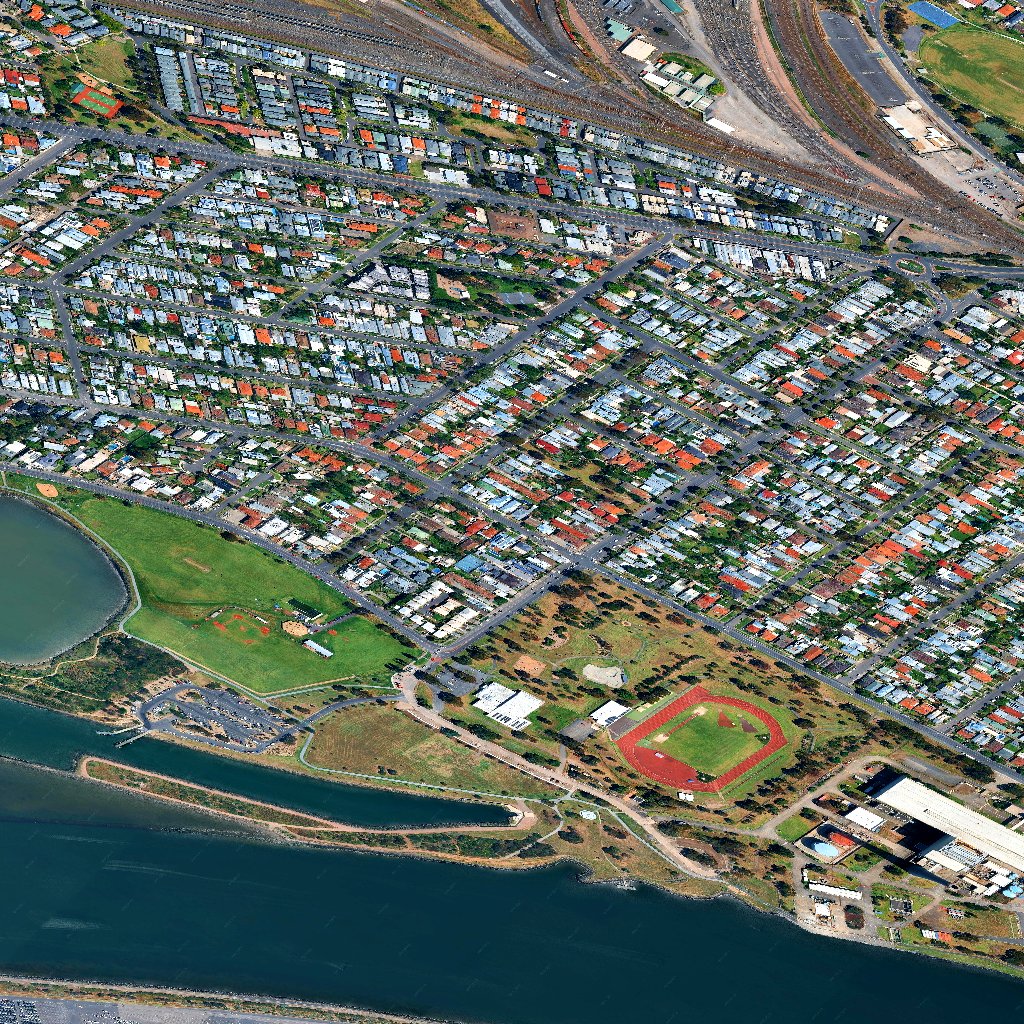}&\hspace{2pt}%
    \includegraphics[width=0.16\textwidth]{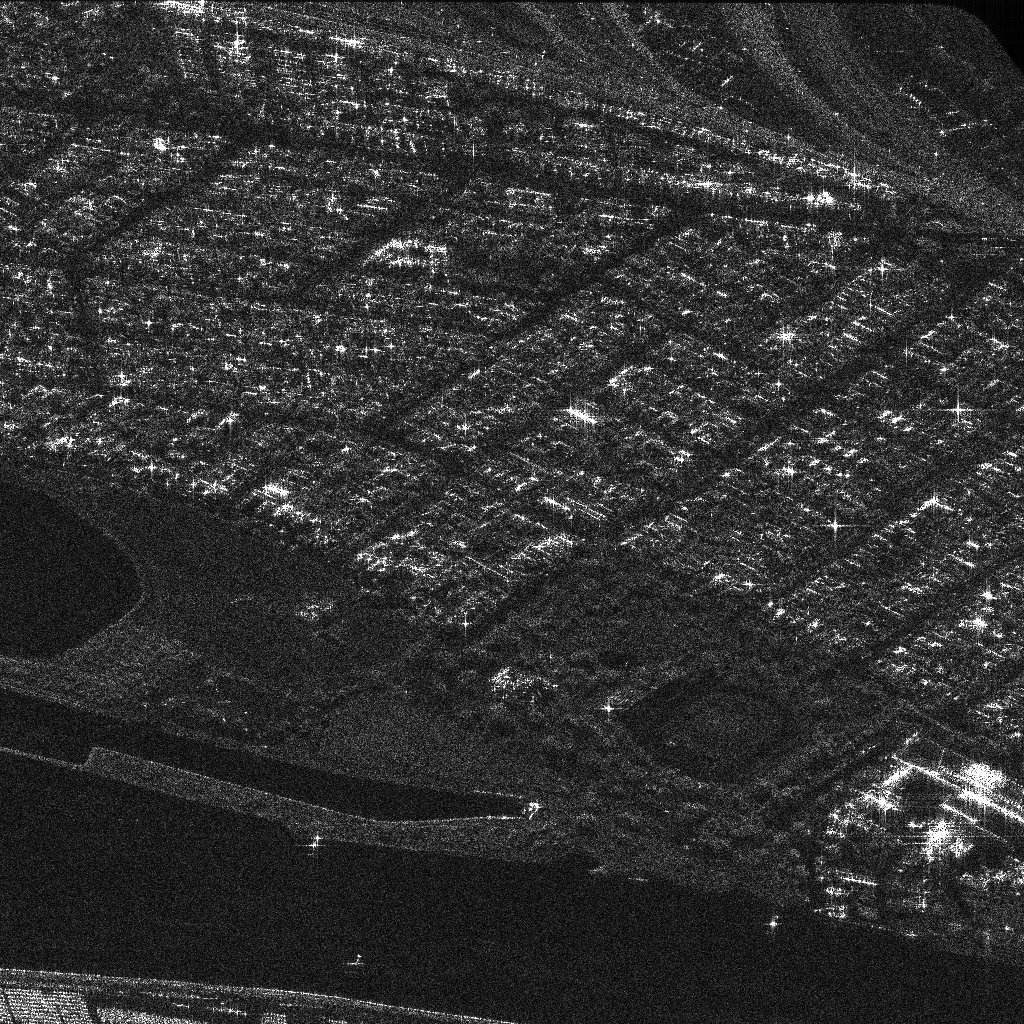}%
  &     \includegraphics[width=0.16\textwidth]{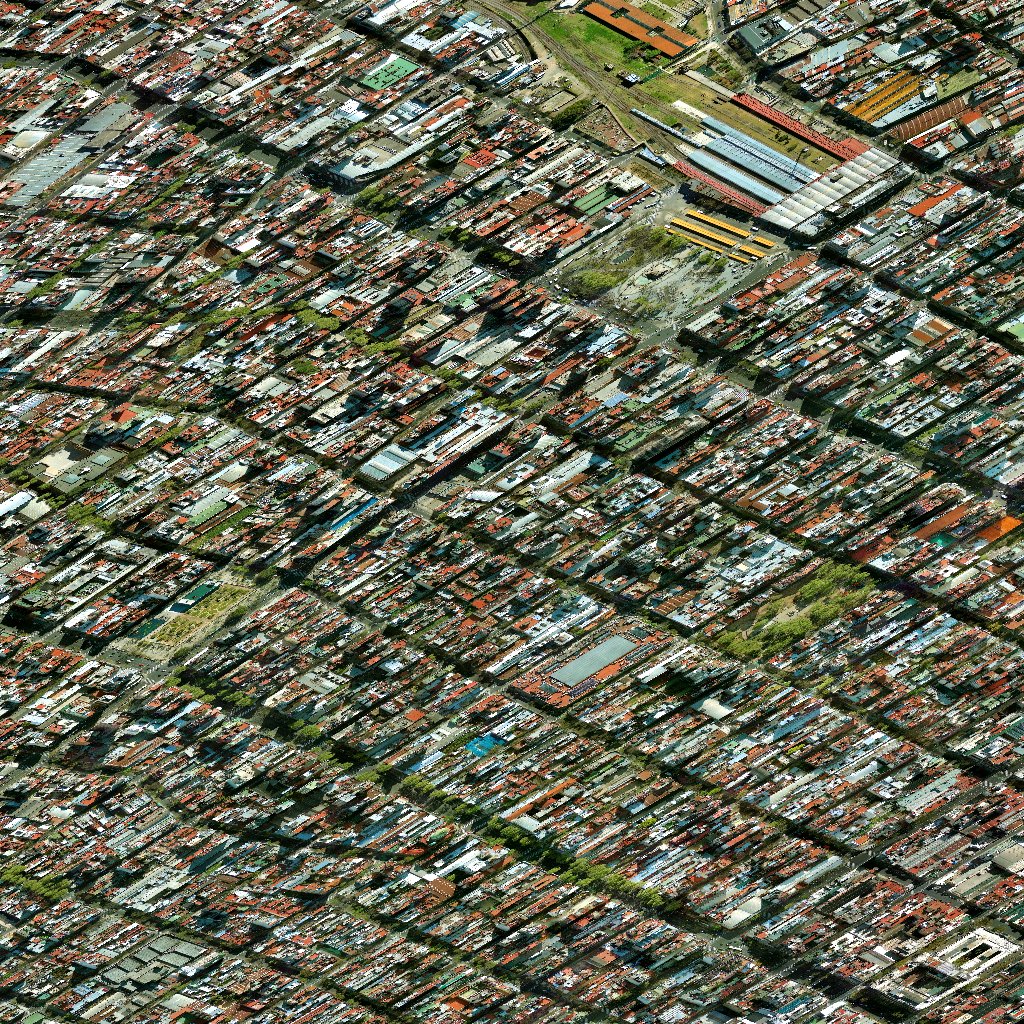}&\hspace{2pt}%
    \includegraphics[width=0.16\textwidth]{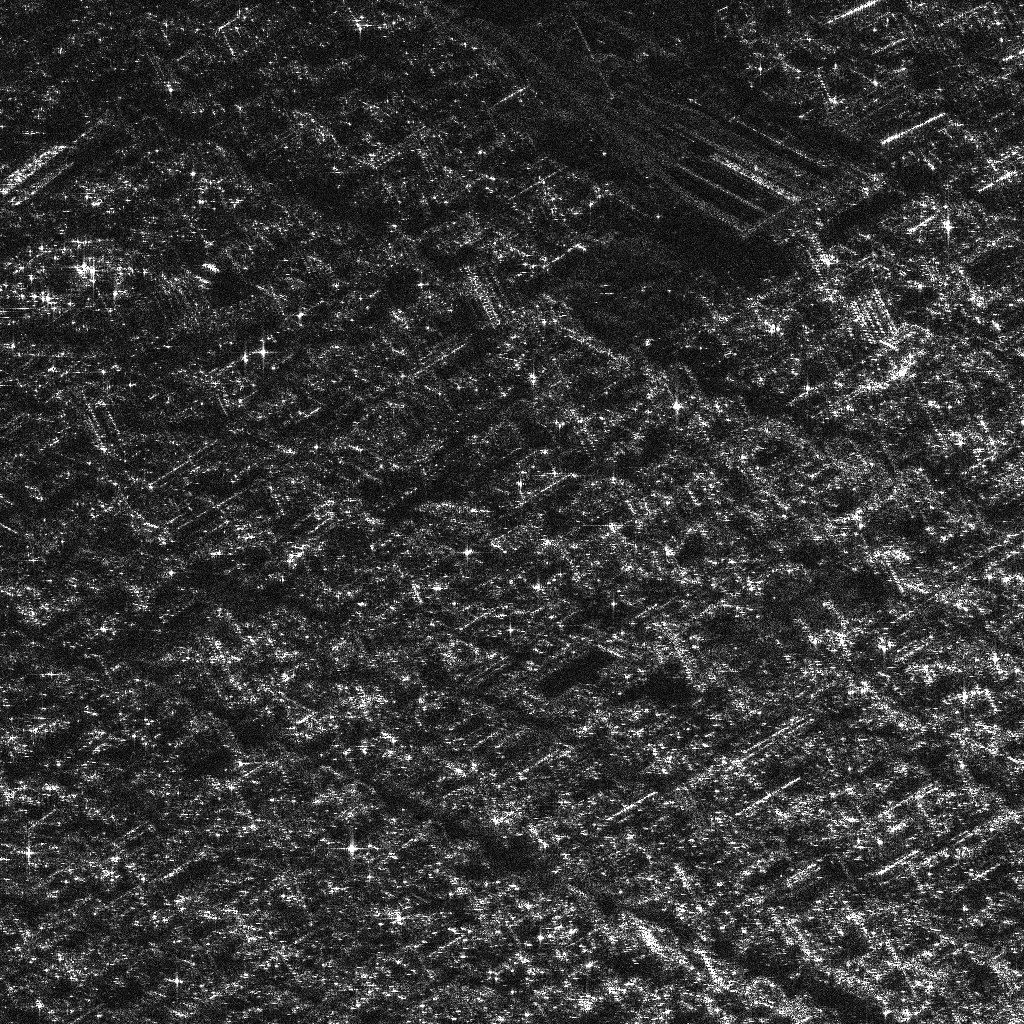}  \\
\multicolumn{2}{c}{\small (d)} & \multicolumn{2}{c}{\small (e)} & \multicolumn{2}{c}{\small (f)} \\[2pt]
    \includegraphics[width=0.16\textwidth]{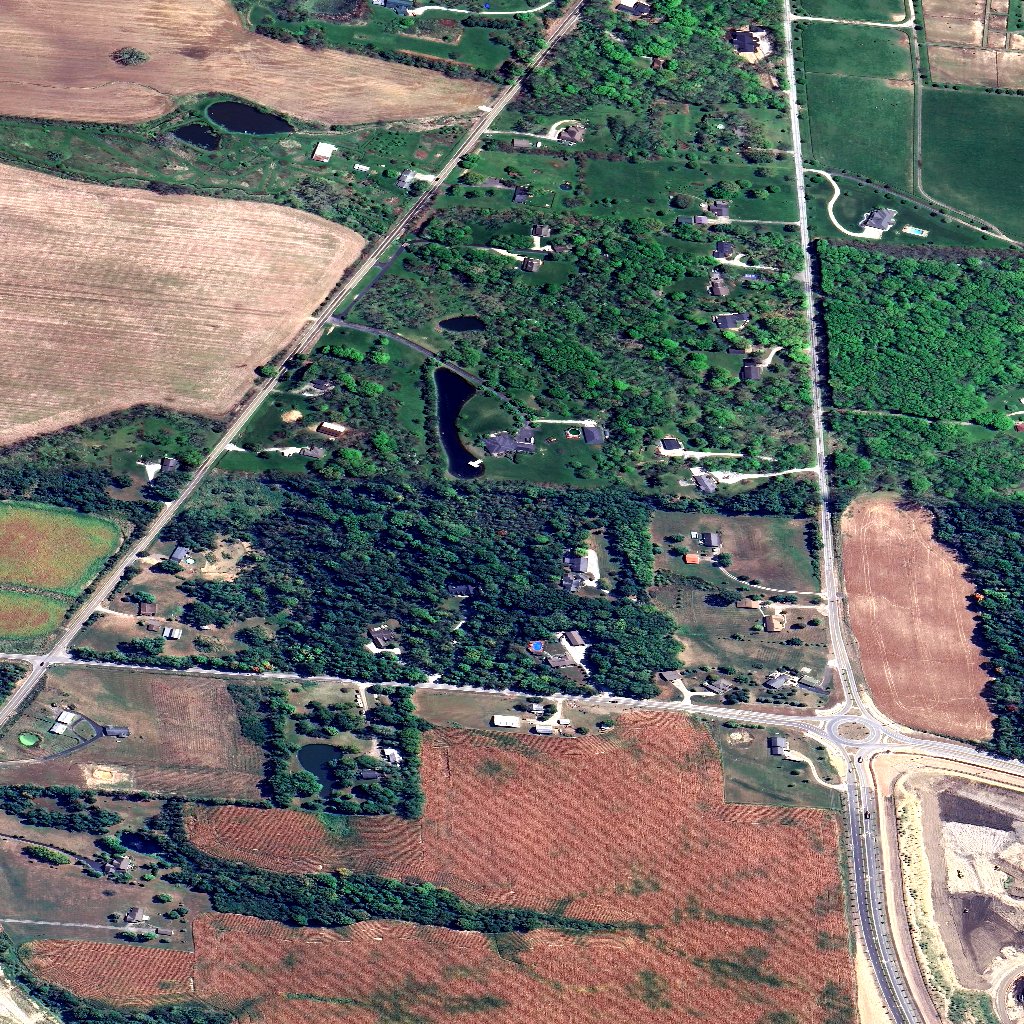}&\hspace{2pt}%
    \includegraphics[width=0.16\textwidth]{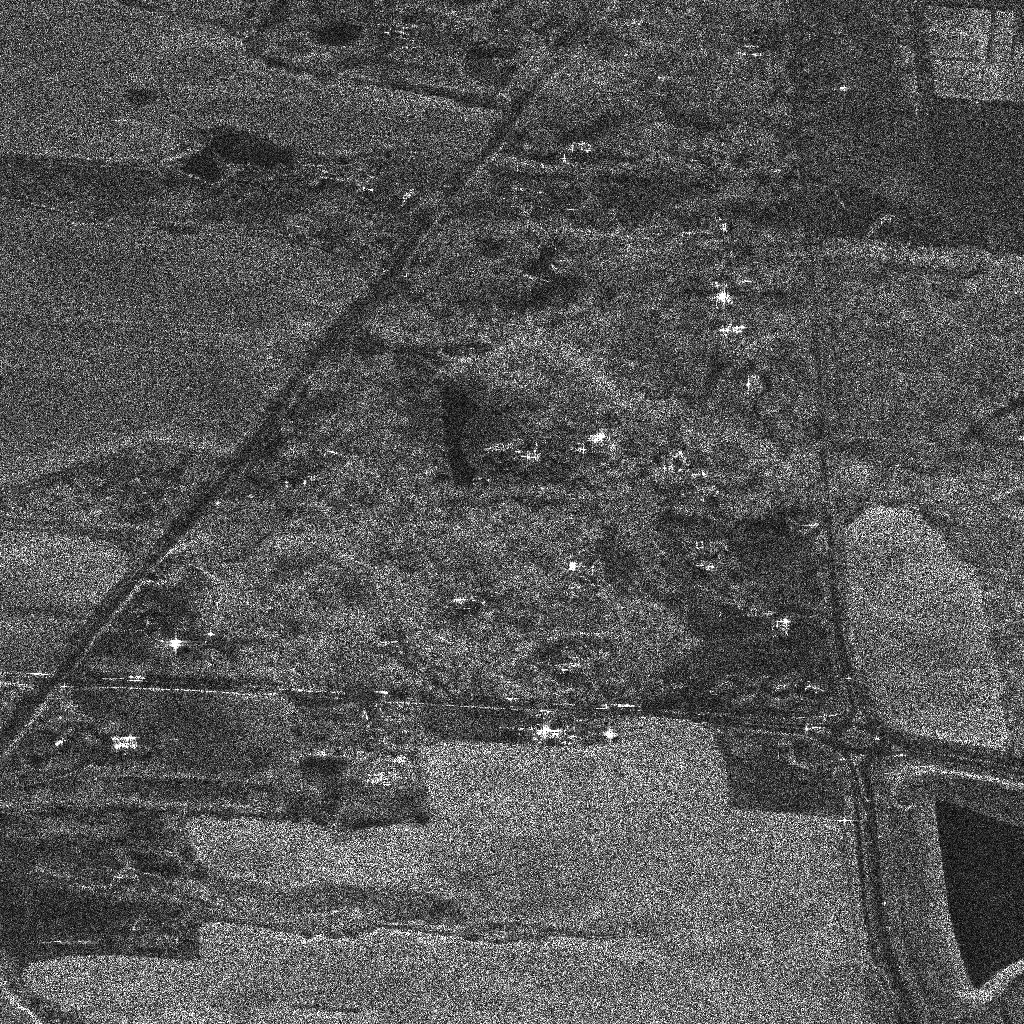}%
  &     \includegraphics[width=0.16\textwidth]{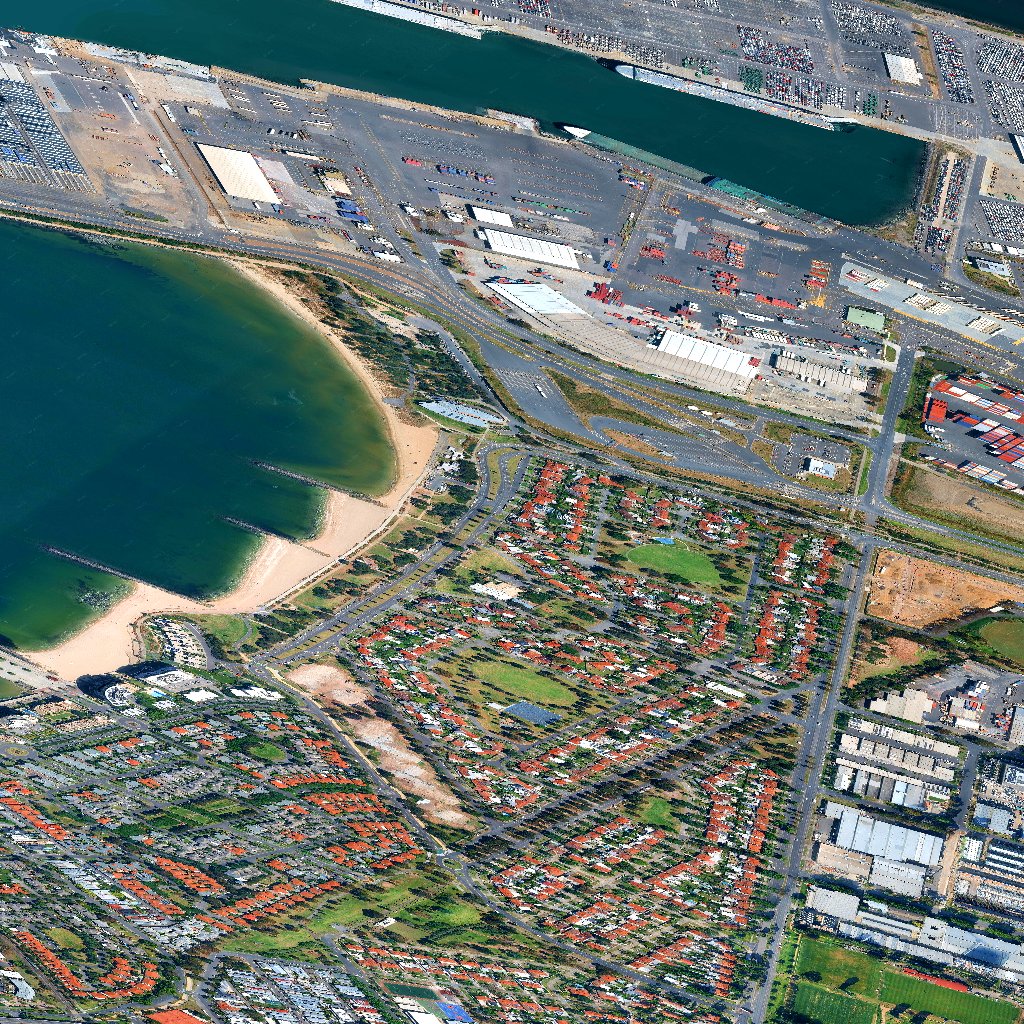}&\hspace{2pt}%
    \includegraphics[width=0.16\textwidth]{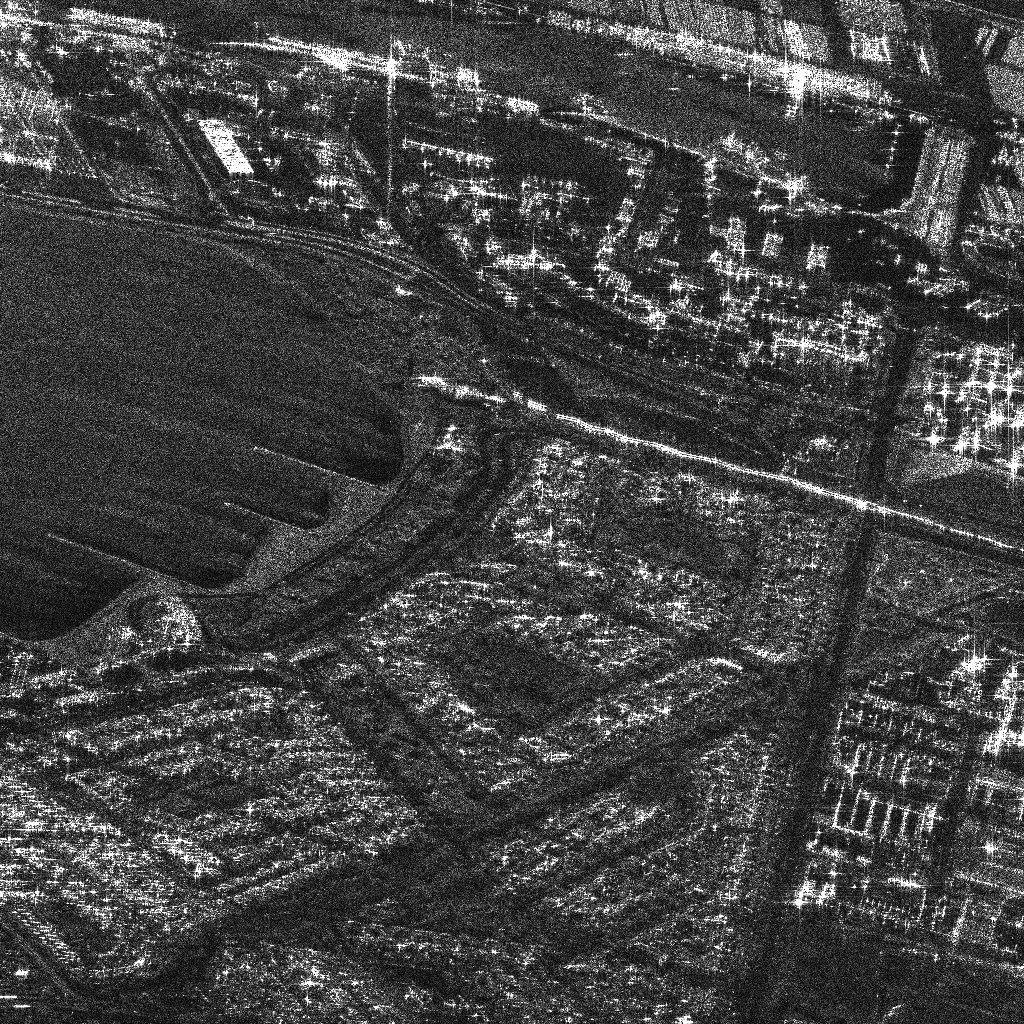}%
  &     \includegraphics[width=0.16\textwidth]{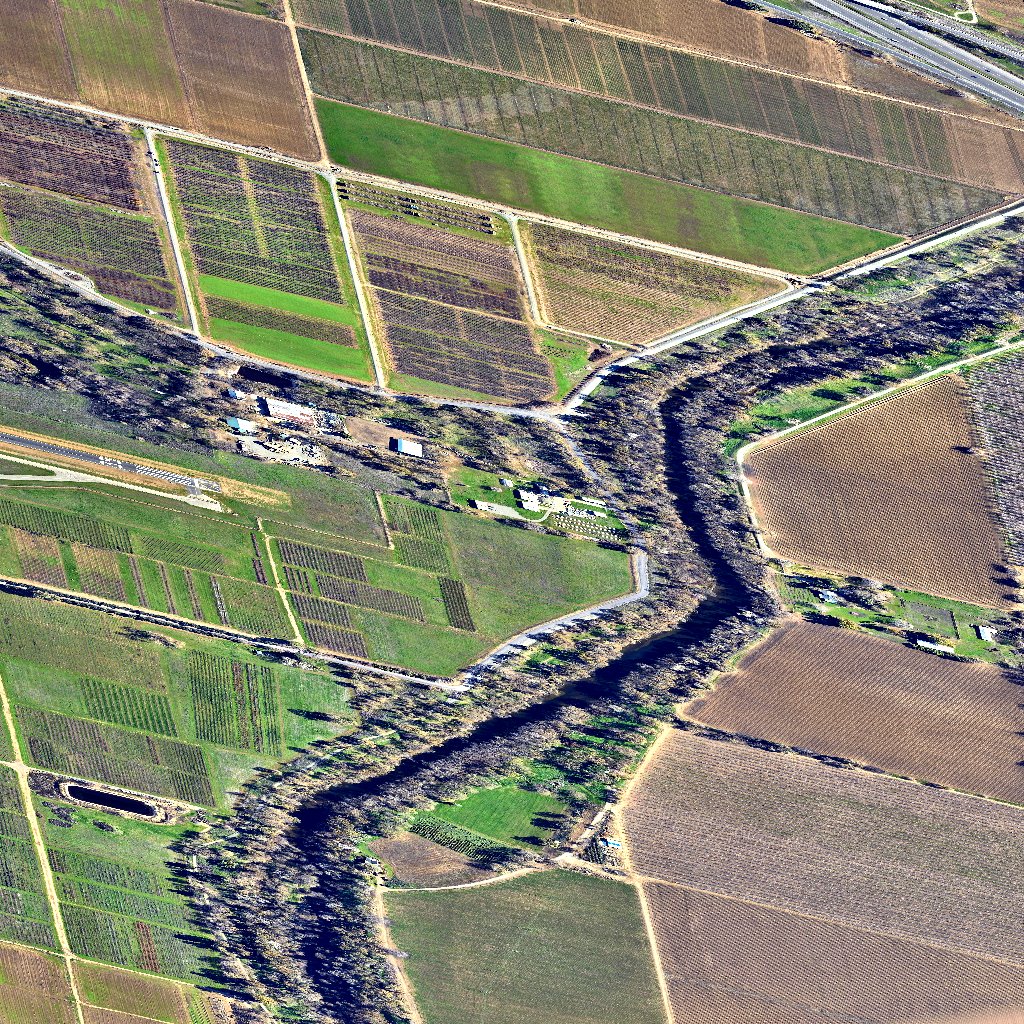}&\hspace{2pt}%
    \includegraphics[width=0.16\textwidth]{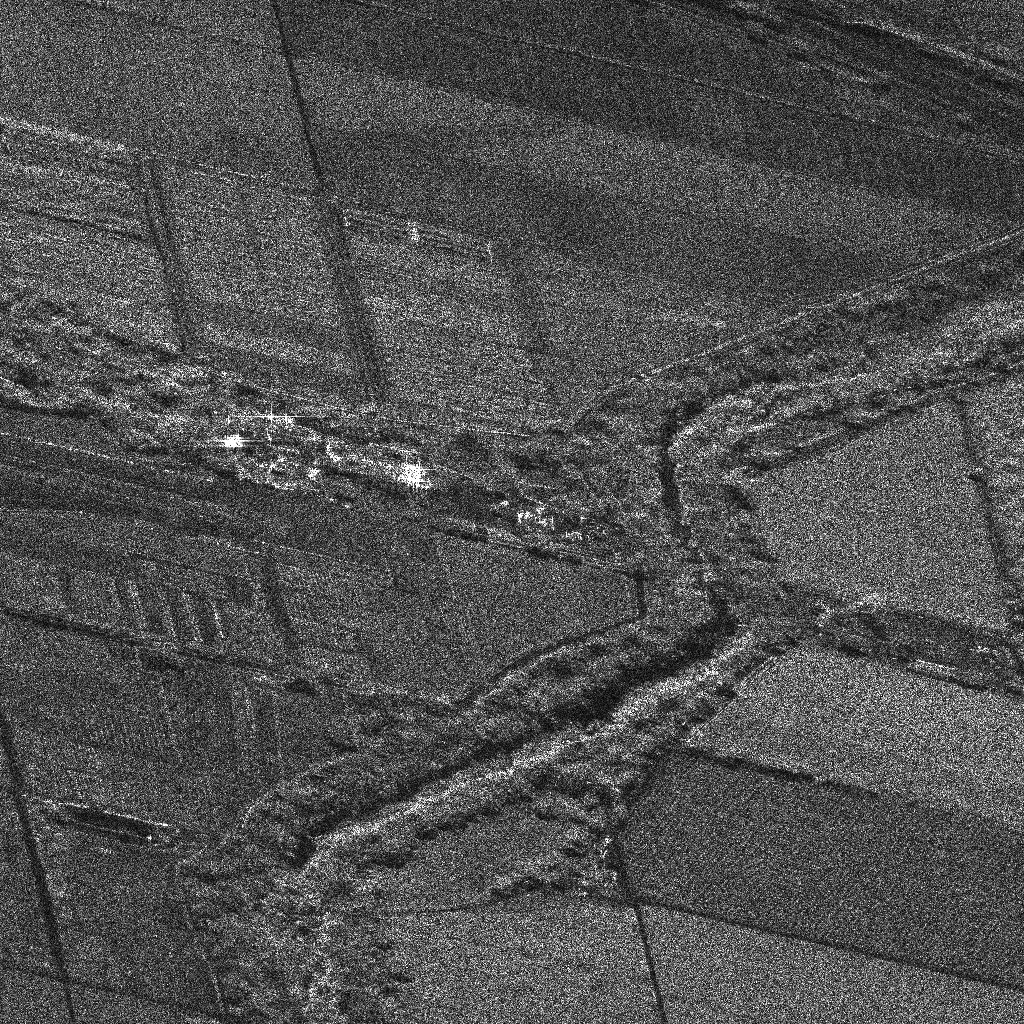}  \\
\multicolumn{2}{c}{\small (g)} & \multicolumn{2}{c}{\small (h)} & \multicolumn{2}{c}{\small (i)} \\[2pt]
    \includegraphics[width=0.16\textwidth]{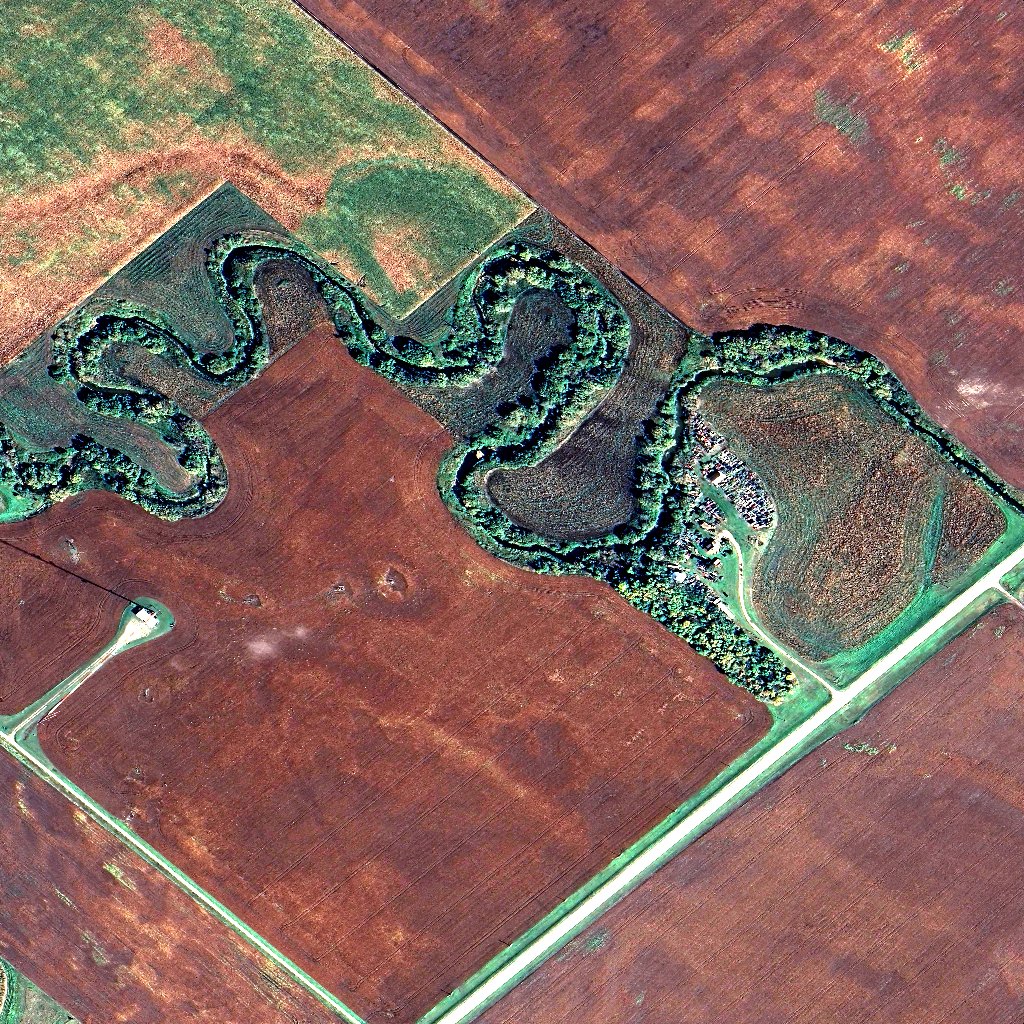}&\hspace{2pt}%
    \includegraphics[width=0.16\textwidth]{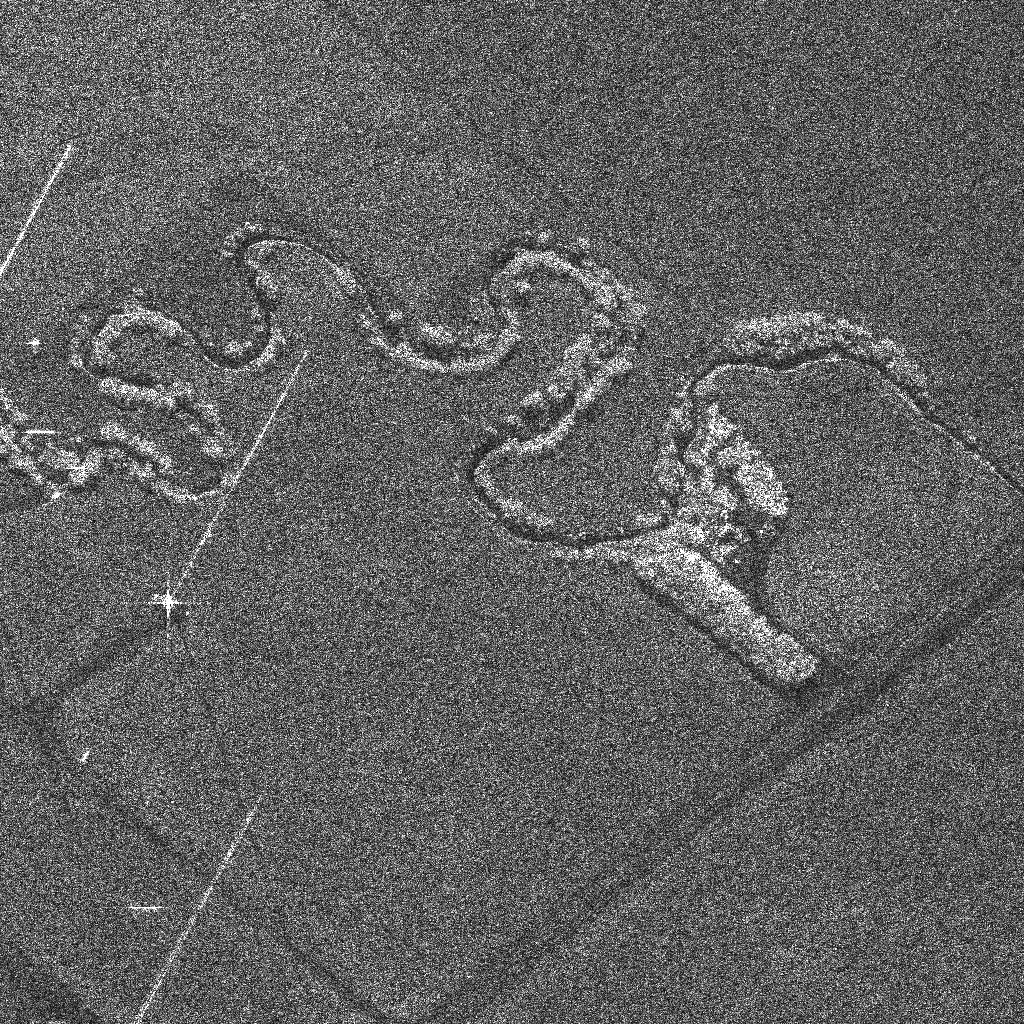}%
  &     \includegraphics[width=0.16\textwidth]{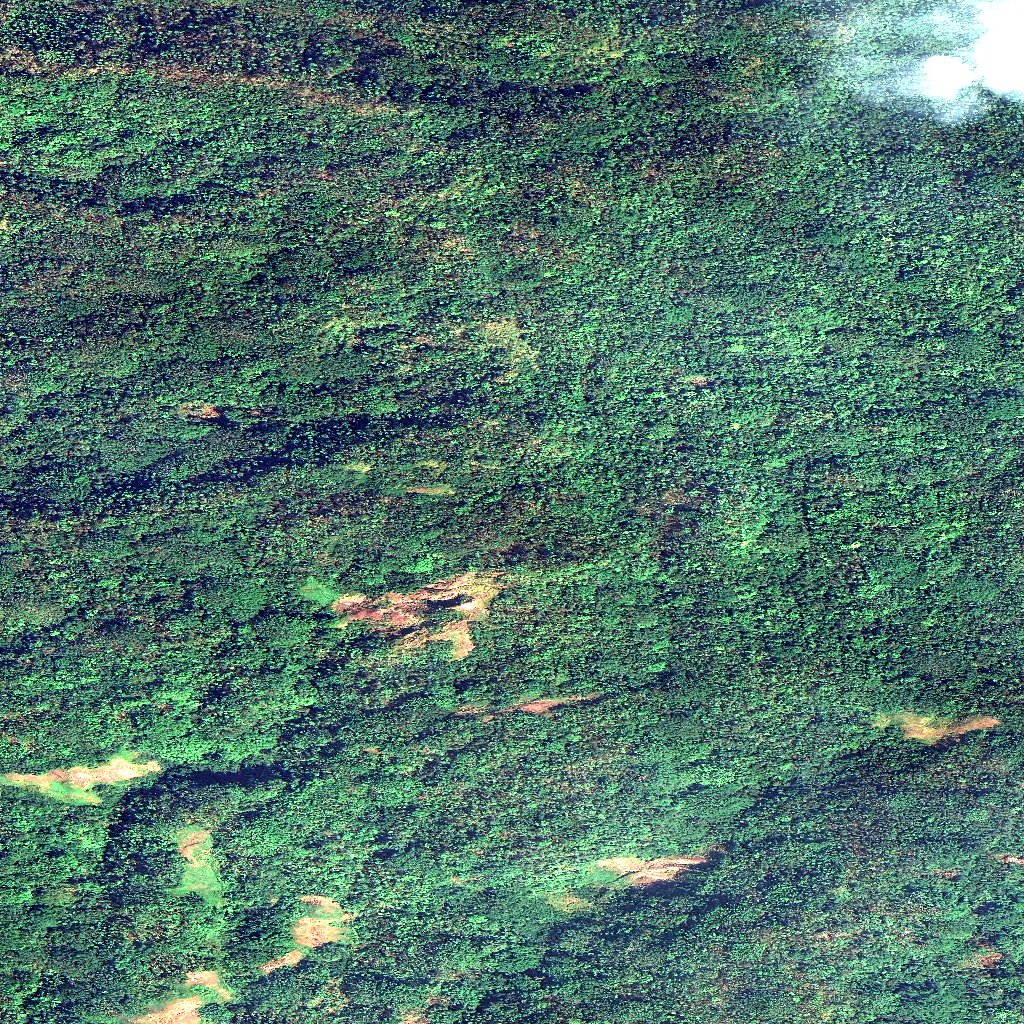}&\hspace{2pt}%
    \includegraphics[width=0.16\textwidth]{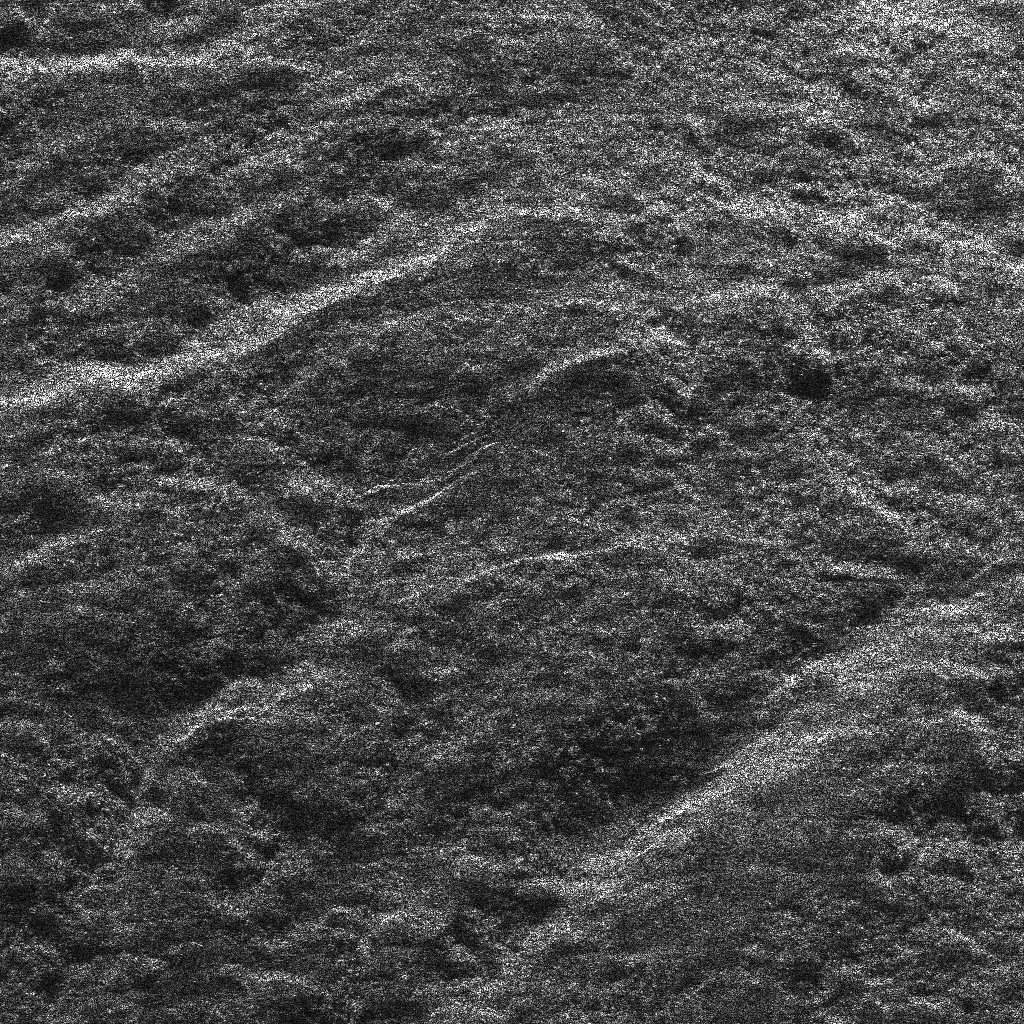}%
  &     \includegraphics[width=0.16\textwidth]{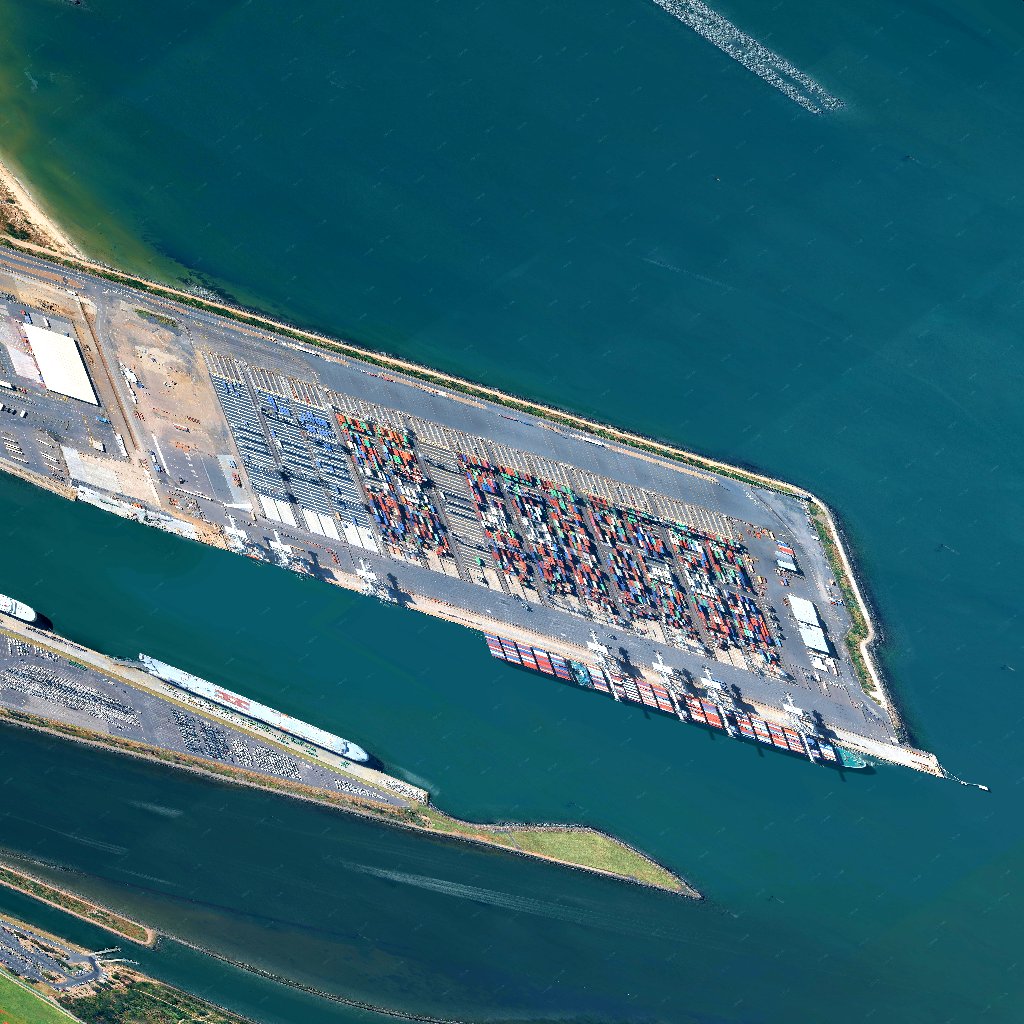}&\hspace{2pt}%
    \includegraphics[width=0.16\textwidth]{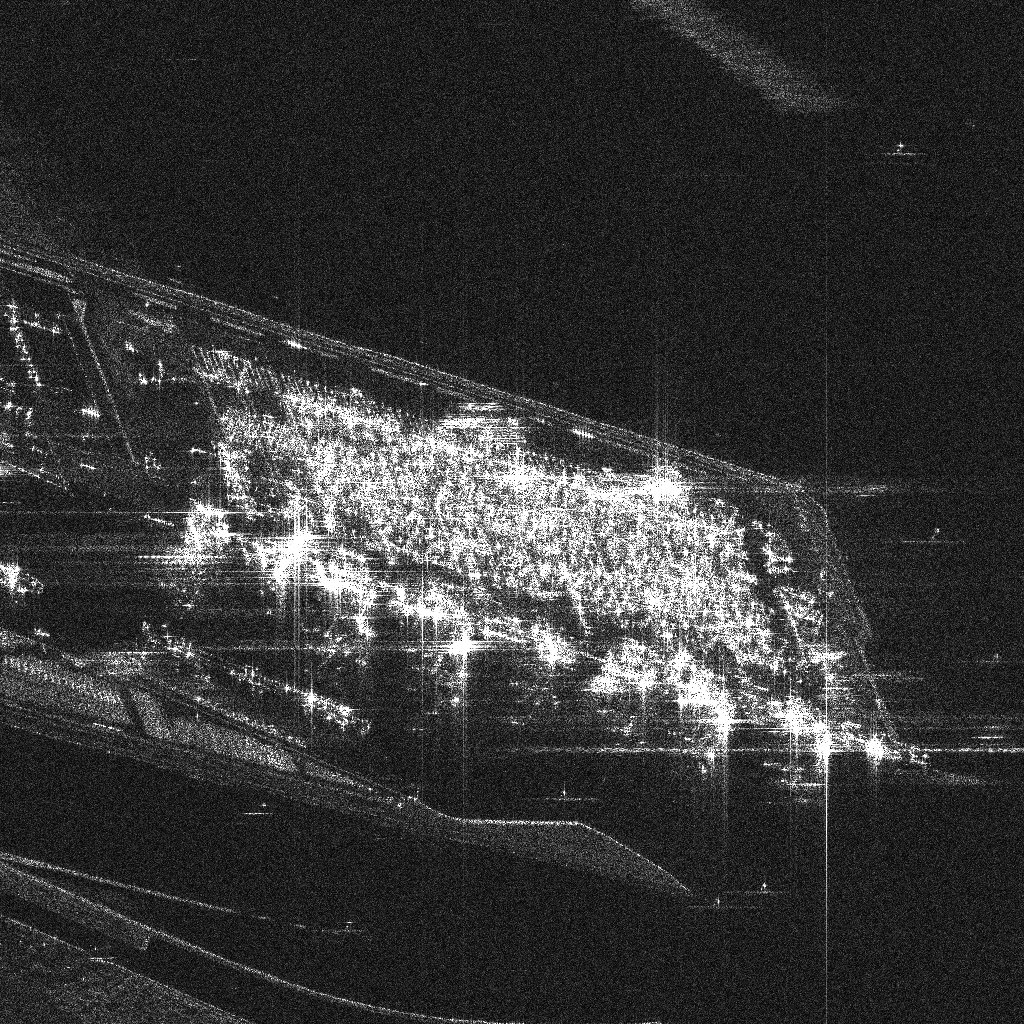}  \\
\multicolumn{2}{c}{\small (j)} & \multicolumn{2}{c}{\small (k)} & \multicolumn{2}{c}{\small (l)} \\[2pt]
\end{tabular}
\end{adjustbox}

\caption{Examples of selected SAR--optical pairs from our dataset.
(a) A satellite image of a coastal region featuring dense buildings, roads, parking lots, and ships docked at a port. Green spaces and a railway track are also visible.
(b) A satellite image of a large highway interchange featuring multiple loops and ramps, surrounded by patches of agricultural land and a large water body, with a few isolated buildings visible.
(c) A satellite image of a vast landscape featuring intricate erosion patterns, winding roads, and clusters of structures, highlighting linear features and geological formations.
(d) A satellite image of a rugged terrain featuring rocky outcrops, sparse vegetation patches, and winding paths, with some darker areas possibly indicating water bodies.
(e) A satellite image of an urban area featuring a grid-like road system, diverse residential buildings, a large sports field, adjacent water bodies, and patches of greenery.
(f) A satellite image of a densely populated urban area featuring tightly packed buildings, diagonal streets, and patches of green space, with a railway track running diagonally through the city.
(g) A satellite image of a landscape featuring agricultural fields with varied crops, residential clusters, roads, and water bodies, with some construction or excavation sites visible.
(h) A satellite image of a coastal region displaying a dense urban area with residential neighborhoods, a large port including multiple docks and shipping containers, green spaces, roads, and a beachfront.
(i) A satellite image of a rural landscape featuring geometrically organized agricultural fields, a winding river, roads, and small buildings.
(j) A satellite image of an area featuring rectangular farmland plots, intersected by roads and a winding water body, with a cluster of structures adjacent to the water.
(k) A satellite image of a dense forest canopy featuring uniform green tones interspersed with patches of lighter shades, likely indicating clearings or water bodies, and a cloud formation on the upper right.
(l) A satellite image of a port featuring neatly arranged shipping containers in a structured grid, a pier extending into water, adjacent roads, and nearby green areas.}
\label{fig:gallery_selected_pairs}
\end{figure*}

%
%
\bibliographystyle{splncs04}
\bibliography{main}
\end{document}